\documentclass{article}

\usepackage{microtype}
\usepackage{graphicx}
\usepackage{subcaption}
\usepackage{booktabs} 

\usepackage{hyperref}

\usepackage{algorithm}

\usepackage{algpseudocode}
\usepackage{multirow}
\usepackage{dashrule}

\usepackage[accepted]{icml2026}

\usepackage{amsmath}
\usepackage{amssymb}
\usepackage{mathtools}
\usepackage{amsthm}
\usepackage{bm}

\renewcommand{\vec}[1]{\mbox{\boldmath${#1}$}}
\newcommand{\vx}{\vec{x}}
\newcommand{\vX}{\vec{X}}
\newcommand{\vphi}{\vec{\phi}}
\newcommand{\vPhi}{\vec{\Phi}}
\newcommand{\vW}{\vec{W}}
\newcommand{\vL}{\vec{L}}
\newcommand{\vD}{\vec{D}}
\newcommand{\vI}{\vec{I}}
\newcommand{\vU}{\vec{U}}
\newcommand{\vV}{\vec{V}}
\newcommand{\vu}{\vec{u}}
\newcommand{\vv}{\vec{v}}
\newcommand{\vTheta}{\vec{\Theta}}
\newcommand{\vGamma}{\vec{\Gamma}}
\newcommand{\vb}{\vec{b}}

\newcommand{\camready}[1]{#1} 

\usepackage[capitalize,noabbrev]{cleveref}

\theoremstyle{plain}

\theoremstyle{definition}

\theoremstyle{remark}

\usepackage[textsize=tiny]{todonotes}

\begin{document}

\twocolumn[
  \icmltitle{Cluster-Aware Causal Mixer for Online Anomaly Detection in Multivariate Time Series}



  \icmlsetsymbol{equal}{*}

  \begin{icmlauthorlist}
    \icmlauthor{Md Mahmuddun Nabi Murad}{USF}
    \icmlauthor{Yasin Yilmaz}{USF}
  \end{icmlauthorlist}
    \icmlaffiliation{USF}{Department of Electrical Engineering, University of South Florida, Tampa, FL, USA}
    \icmlcorrespondingauthor{Md Mahmuddun Nabi Murad}{mmurad@usf.edu}


  \icmlkeywords{Machine Learning, ICML}

  \vskip 0.3in
]



\printAffiliationsAndNotice{}  


\begin{abstract}
Early and accurate detection of anomalies in time-series data is critical due to the substantial risks associated with false or missed detections. While MLP-based mixer models have shown promise in time-series analysis, they do not maintain temporal causality during data processing. Moreover, real-world multivariate time series often contain numerous channels with diverse inter-channel correlations. Spurious correlations in the reconstructed time series lead to noisy representations, resulting in inaccurate anomaly detection. In addition, anomaly scoring methods that ignore temporal continuity can mislead sequential detection. To address these challenges, we propose a cluster-aware causal mixer for multivariate time-series anomaly detection. Channels are grouped into clusters based on their correlations, and each cluster is embedded through a dedicated embedding layer. A causal mixer is introduced to integrate information while maintaining temporal causality. We further develop a sequential anomaly-scoring method that accumulates evidence over time and refines anomaly boundaries. Our proposed model operates in an online fashion, making it suitable for real-time time-series anomaly detection. Experimental evaluations across six public benchmark datasets demonstrate that the proposed approach consistently achieves superior performance.
\end{abstract}

\section{Introduction}
Multivariate time series data consists of sequential measurements collected from multiple sources (e.g., sensors) over time. When there is an anomaly such as sensor malfunction or malicious data manipulation, the resulting patterns often deviate from normal behavior. Accurate and timely detection of such anomalies is critical in a wide range of applications, including the identification of cyberattacks~\cite{cyber} or sensor failures in critical infrastructure systems, such as water distribution networks~\cite{swat}. Considering the open-set possibilities for anomalies and the difficulty to sufficiently label and train on anomalous samples in real-world datasets, most research in time series anomaly detection has focused on unsupervised methods~\cite{anom_review}. These approaches generally involve training models to learn the temporal (and spatial for multivariate analysis) characteristics of normal training data.\par


The mainstream deep learning approaches learn normal patterns through training prediction or reconstruction models, in which anomalous test instances are expected to cause statistically larger forecast or reconstruction errors. Recent methods further incorporate uncertainty estimation, assigning low uncertainty to normal data and higher uncertainty to anomalous observations~\cite{uncertainty}.\par
In recent years, transformer-based models have gained popularity in time series analysis, with many studies adopting transformer variants as the core architecture~\cite{gta, anomaly_transformer, d3r, tranad, npsr, sensitivehue}. Among them, NPSR~\cite{npsr} and SensitiveHUE~\cite{sensitivehue} further enhance performance by refining anomaly scoring mechanisms. In addition, recent studies, including~\cite{zeng2023transformers}, question the effectiveness of transformer-based models for time series forecasting, showing that simpler multilayer perceptron (MLP) based models can achieve comparable performance. Furthermore, MLP-Mixer models have shown superior performance against transformer variants in time series forecasting tasks~\cite{wpmixer, tsmixer}. 

While transformers enforce temporal causality via attention masking, existing MLP-Mixer models~\cite{patchad} do not have such mechanisms. This motivates us to \emph{study whether an explicit mechanism to ensure causal mixing can help in detecting anomalies in multivariate time series.}\\

Our contributions are threefold: 
\begin{itemize} 
\item \textbf{First}, motivated by the strong performance of MLP-Mixer models and addressing their limitations regarding causality, we propose a novel \textbf{causal mixer} module that enforces strict temporal causality, ensuring that each representation at time $t$ depends only on past and present information, eliminating any future leakage.

\item \textbf{Second}, to capture meaningful inter-channel dependencies and mitigate spurious correlations in normal data, we introduce \textbf{cluster-aware multi-embedding}, which enhances the discriminability between normal and anomalous representations.

\item \textbf{Third}, we propose a \textbf{sequential anomaly scoring method} that accumulates anomaly evidence over time, refining anomaly segment boundaries and improving overall anomaly detection performance.
\end{itemize}

\section{Related Work}
Time series anomaly detection methods are commonly classified as unsupervised~\cite{munir2018deepant} or supervised~\cite{ma2016supervised}, and further categorized into point-based~\cite{point_anom} or sequence-based~\cite{doshi2022reward} approaches. \camready{Extensive research on time-series anomaly detection has also been conducted in statistics through change-point detection and outlier detection methods~\cite{aminikhanghahi2017survey}. Classical machine learning approaches, such as k-nearest-neighbors (kNN), support vector machine (SVM), and Isolation Forest, have also been applied for anomaly detection~\cite{chandola2009anomaly}. With the reduced need for manual feature engineering and the increasing availability of computational resources, deep learning methods have become increasingly popular for time-series anomaly detection.}
With the rise of deep learning, models like LSTM~\cite{lstm_7, oracleAD} have been used to capture long-term dependencies, and hybrid approaches such as LSTM-VAE have emerged~\cite{lstm_vae}. Autoencoder-based models learn low-dimensional representations of normal data~\cite{dagmm, uae, usad}. Time series can also be represented as graphs, enabling graph-based models like GDN~\cite{gdn}, MTAD-GAT~\cite{mtad_gat}, and Graph-MoE~\cite{huang2025graph}. Recently, transformer-based models have shown superior performance in anomaly detection~\cite{gta, anomaly_transformer, d3r, tranad,npsr, sensitivehue}. Some models improve the anomaly detection performance by refining the anomaly scores~\cite{npsr, sensitivehue, sat}.\par

NPSR~\cite{npsr} introduces a nominality score that accounts for the influence of neighboring points while SensitiveHUE~\cite{sensitivehue} incorporates heteroscedastic uncertainty into the reconstruction loss. To capture spatio-temporal dependencies, SensitiveHUE employs statistical feature elimination and robust normalization based on the median and interquartile range of anomaly scores, derived from the entire test set. However, this normalization requires access to the anomaly score of the entire test set beforehand, limiting the model’s applicability in online anomaly detection. Although removing the robust normalization enables real-time detection, it substantially reduces performance, highlighting a key limitation of SensitiveHUE in scenarios requiring timely anomaly detection.\par

Recent studies~\cite{position} show that simple baselines, such as PCA\_Error, can outperform or perform similarly to complex models with Transformer and Graph-based variants, suggesting a simple architecture for the reconstruction model. Furthermore, the performance of the transformer-based models is questioned in paper~\cite{zeng2023transformers}, while MLP-mixer-based models~\cite{wpmixer, timemixer, tsmixer, tiny_tsmixer, patch_mixer, kanmixer} outperform transformer variants in time series analysis. To ensure causality in time series, transformers employ masking in attention. However, to the best of our knowledge, no existing MLP-Mixer model incorporates mechanisms to preserve temporal causality during the mixing process.\par
To address these limitations, we propose a novel \textbf{C}luster-aware \textbf{C}ausal \textbf{M}ixer model for \textbf{T}ime Series \textbf{A}nomaly \textbf{D}etection (CCM-TAD). CCM-TAD enforces temporal causality within MLP-Mixer layers, ensuring that each time step’s representation depends only on past and present inputs. Unlike causal discovery methods, our objective is anomaly detection with causally correct temporal mixing, enhanced by cluster-aware multi-embedding to improve normal data modeling and a sequential statistical scoring method with refined anomaly boundaries.

\section{Proposed Method}
\label{proposed_method}

Let \( \mathcal{X}=\{\vx_1,\vx_2 \dots, \vx_t, \dots\} \) represent a multivariate time series, where \( \vx_t \in \mathbb{R}^{1 \times C} \) represents the observation at time $t$, and $C$ is the number of the channels. Each instance $\vx_t$ is associated with a label \( y_t \in \{0, 1\} \), where \( y_t = 1 \) and \( y_t = 0 \) indicate anomalous and normal states, respectively. The goal is to predict $y_t$ for each time step. \par
The training set comprises only normal instances, while the test set includes both normal and anomalous data. At each time step \( t \), a look-back window  $\vX_L = \{\vx_{t-L+1}, \dots, \vx_t\} \in \mathbb{R}^{L \times C}$ is input to the model to reconstruct the latest observation \( \vx_t \). The reconstruction loss between \( \vx_t \) and the reconstructed \( \hat{\vx}_t \) is then used to derive the anomaly score. For subsequent predictions, the look-back window moves forward by one time step, and the process repeats iteratively. All data are normalized using min–max scaling, with statistics computed from the training set. \par
\camready{Note that, similar to prior reconstruction-based anomaly detection methods, we assume that the training data is predominantly clean. Although this setting is commonly referred to as unsupervised anomaly detection in the literature, contamination in the training set may degrade detection performance.}

\begin{figure*}[bt!]
    \centering
    \includegraphics[width=\textwidth, height=0.33\textheight, keepaspectratio]{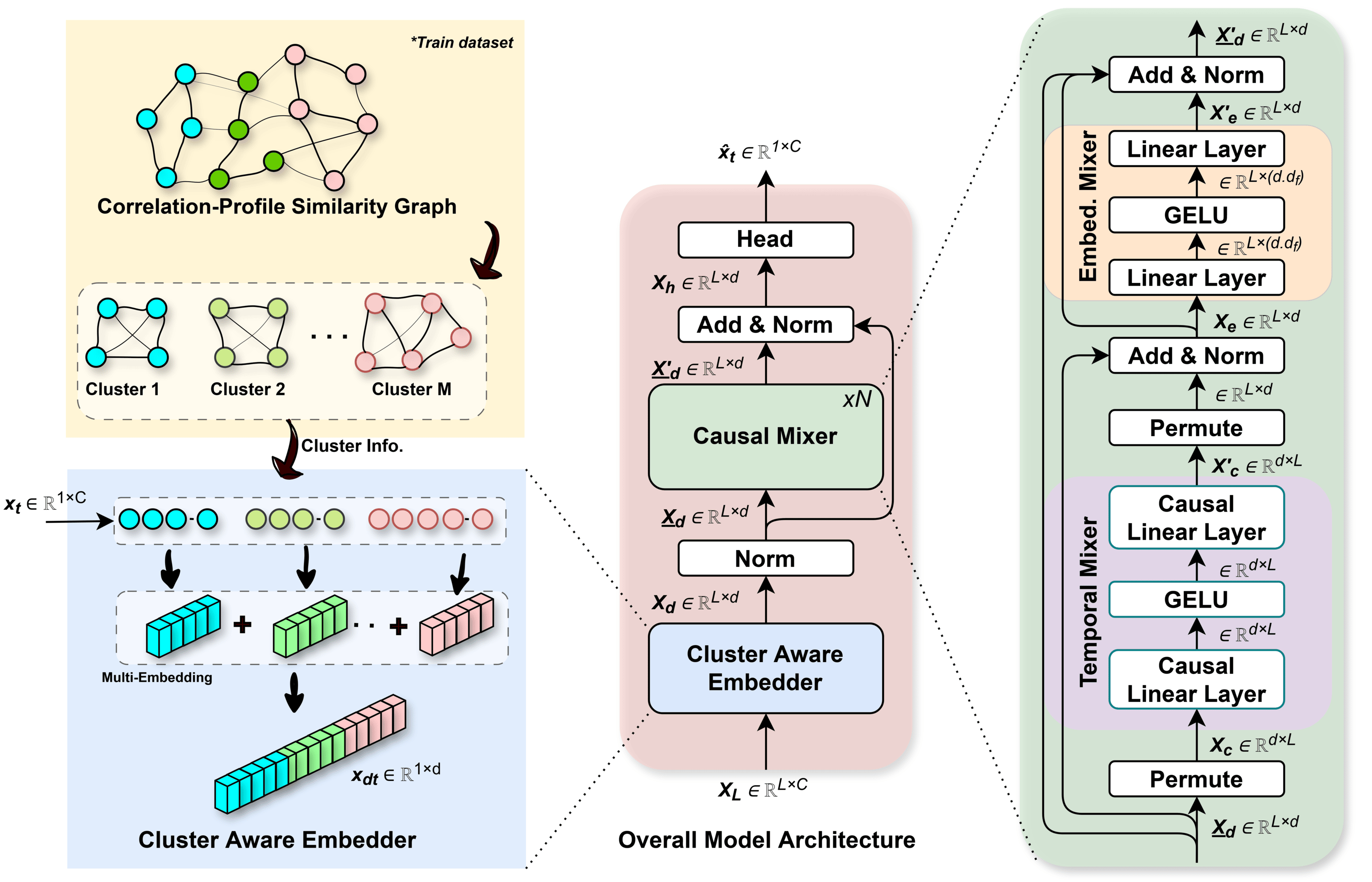}
    \caption{An architectural view of our reconstruction model. First, the multivariate input is embedded into a $d$-dimensional representation using a cluster-aware embedding module. Then, a causal mixer is used to extract features for reconstruction. In the correlation-profile similarity graph, each node represents a channel, and edge weights correspond to the cosine similarity between its correlation profile and those of its neighbors. \camready{Code is available at: \url{https://github.com/Secure-and-Intelligent-Systems-Lab/CCM-TAD}}}
    \label{fig:overall_model_architecture}
    \vspace{-3mm}
\end{figure*}

\subsection{Overall Model Architecture}
\label{model_architecture}
The reconstruction architecture of our Cluster-aware Causal Mixer model for Time Series Anomaly Detection (CCM-TAD) is depicted in Figure \ref{fig:overall_model_architecture}. Our model consists of two key components: cluster-aware multi-embedding and causal mixer. At the beginning of our model, we pass the input \(\vX_L \in \mathbb{R}^{L \times C}\) through the cluster-aware multi-embedding module (Section~\ref{cluster_aware_multi_embedding}) to get embedded output \(\vX_d \in \mathbb{R}^{L \times d}\). 
We then apply 1D batch normalization to the output of the embedding module, resulting in \(\underline{\vX_d} \in \mathbb{R}^{L \times d}\). \(\underline{\vX_d}\) is then processed through multiple causal mixer modules. Each causal mixer module (Section \ref{causal_mixer}) comprises a temporal mixer module and an embedding mixer module. The temporal mixer module consists of two causal linear layers connected by a non-linear GELU activation. The operations of the temporal mixer module are summarized as
\begin{flalign}
    & \vX_c = \mathcal{P}(\underline{\vX_d}) \in \mathbb{R}^{d \times L} \\
    & \vX^{'}_c = \mathcal{L}_2(\mathcal{G}(\mathcal{L}_1(\vX_c))) \in \mathbb{R}^{d \times L} \\
    & \vX_e = \mathcal{BN}(\mathcal{P}(\vX^{'}_c) + \underline{\vX_d})) \in \mathbb{R}^{L \times d},
\end{flalign}
where $\mathcal{L}_1: \mathbb{R}^{\cdots\times L} \rightarrow \mathbb{R}^{\cdots\times L}$, $\mathcal{L}_2: \mathbb{R}^{\cdots\times L} \rightarrow \mathbb{R}^{\cdots\times L}$, $\mathcal{G}(.)$, $\mathcal{P}(.)$, and $\mathcal{BN}(.)$ represent the causal linear layer-1, causal linear layer-2, GELU activation, permute operation, and 1D batch normalization, respectively. After the temporal mixer, the data passes through the embedding mixer module: 
\begin{flalign}
    & \vX^{'}_e = \mathcal{L}^{'}_2(\mathcal{G}(\mathcal{L}^{'}_1(\vX_e))) \in \mathbb{R}^{L \times d} \\
    & \underline{\vX^{'}_{d}} = \mathcal{BN}(\vX^{'}_e + \vX_e + \underline{\vX_d}) \in \mathbb{R}^{L \times d}
\end{flalign}

where, $\mathcal{L}^{'}_1: \mathbb{R}^{\cdots\times d} \rightarrow \mathbb{R}^{\cdots\times (d\cdot d_f)}$ and $\mathcal{L}^{'}_2: \mathbb{R}^{\cdots\times (d\cdot d_f)} \rightarrow \mathbb{R}^{\cdots\times d}$ represent standard linear layers, \(d_f\) is a hyperparameter (expansion factor).

After the data is processed through a series of causal mixer modules, it is combined with \(\underline{\vX_d}\) and subsequently normalized, yielding \(\vX_h \in \mathbb{R}^{L \times d}\). This representation serves as the input to the head layer, which consists of a single linear layer $\mathcal{L}_h: \mathbb{R}^{\cdots\times d} \rightarrow \mathbb{R}^{\cdots\times C}$, mapping the \(d\) dimensional data to \(C\) dimensional representation. The model then outputs the last reconstructed observation \(\hat{\vx}_t \in \mathbb{R}^{1 \times C}\), which is further processed to obtain the anomaly score using our proposed anomaly detection method described in~\cref{anomaly_detction_method}.

\subsection{Cluster-Aware Multi-Embedding}
\label{cluster_aware_multi_embedding}
Our proposed cluster-aware multi-embedding combines channel clustering and multi-embedding to enhance normal representation. We first assign each channel a cluster index based on its correlation pattern computed offline from the training dataset. This cluster index for each channel is then utilized in the clustering module during training and inference to group the input channels into clusters. Then, a separate embedding layer is employed for each channel cluster. The outputs of all embedding layers are concatenated to obtain the final high-dimensional representation. This cluster-aware multi-embedding technique mitigates spurious correlations that arise when channels with heterogeneous relationships are jointly embedded, and improves the discriminability between normal and anomalous representations.

\subsubsection{Channel Clustering}
\label{feature_clustering}
Real-world time series data often contains a mix of highly correlated, weakly correlated, and uncorrelated channels. A single shared embedding layer cannot distinguish these relations and may inadvertently impose artificial dependencies during the learning process. 
This can lead to degraded generalization performance, especially in anomaly detection tasks that require precise modeling of normal patterns. To address this, we propose a cluster-aware multi-embedding technique that utilizes spectral clustering~\cite{von2007tutorial} to cluster the channels into \(M\) clusters based on their correlations and assigns a dedicated embedding layer to each group. 

Let \(\vPhi \in [-1,1]^{C \times C}\) denote the Pearson correlation matrix derived from the training data, and \(\vPhi_{\text{abs}} \in [0,1]^{C \times C}\) represent its element-wise absolute value. Channel $i$'s correlations with other channels, given by the $i$th row $\vphi_i \in [0,1]^C$ of $\vPhi_{\text{abs}}$, is used as its correlation profile. Some of the channels in the time series may remain constant throughout the series, resulting in undefined (NaN) correlation values with other channels. To address this, we replace all NaN entries in \(\vPhi\) with zeros before proceeding. During the initial clustering phase, we group all channels with zero-valued correlation profiles in the \(M\)th cluster. Let the number of remaining channels be \(C^{'}\) and the corresponding absolute correlation matrix be \(\vPhi^{'}_{\text{abs}} \in [0,1]^{C^{'} \times C^{'}}\) with rows $\vphi_i^{'}$.

To group \(C^{'}\) channels into $M^{'}=M-1$ clusters based on their correlation profiles, we construct a similarity-based weighted adjacency matrix \(\vW \in [0,1]^{C^{'} \times C^{'}}\), where each off-diagonal entry \(w_{ij}\) represents the cosine similarity between the correlation profiles of channels \(i\) and \(j\) and is given by,
\begin{align}
w_{ij} =
\begin{cases}
\label{eq:cosine_similarity}
\frac{(\phi_i^{'})^T \phi_j^{'}}{\left\| \phi_i^{'}\right\| \left\| \phi_j^{'} \right\|}, & \text{if } i \neq j \\
0, & \text{if } i = j.
\end{cases}
\end{align}
Here, \(\left\|.\right\|\) denotes the Euclidean norm. Next, we compute the normalized graph Laplacian matrix \(\vL_{n}\) as,
\begin{align}
    \vL_{n} = \vI - \vD^{-1/2} \vW \vD^{-1/2}
\end{align}
where, \(\vI\) is the identity matrix and the diagonal matrix \(\vD\) is the degree matrix with diagonal element \(d_{ii} = \sum_j w_{ij}\). We then perform spectral embedding by extracting the first \(M^{'}\) non-trivial eigenvectors of \(\vL_{n}\), corresponding to the smallest non-zero eigenvalues.
Let \(\vV \in \mathbb{R}^{C^{'} \times M^{'}}\) contains these eigenvectors as columns. To normalize the embedded representations, we compute,
\begin{align}
    \vU = \vD^{-1/2} \vV 
\end{align}
where the \(i\)th row of \(\vU\), given by \(\vu_i = d_{ii} ^ {-1/2}  \vv_i \in \mathbb{R}^{M^{'}}\), represents the spectral embedding of channel $i$. Finally, K-Means clustering is applied to the rows of \(\vU\) to cluster the channel embeddings into \(M^{'}\) clusters and get the cluster index for each channel as,
\begin{align}
k^{'} = \textsc{K-Means}(\vU,\ M^{'}). 
\end{align}

A pseudocode for the spectral clustering algorithm is given in Algorithm~\ref{algo:channel_clustering} in Appendix~\ref{app:cluster}.
Also, a detailed discussion of the proposed clustering technique is given in Appendix \ref{Additional_Ablation_Studies}.

\subsubsection{Multi-Embedding Layer}
\label{multi_embedding}
After obtaining the cluster indices based on the training dataset, we construct \(M\) clusters of channels for both the training and the test datasets. 
We employ multiple embedders, one for each cluster, to learn channel representations in a cluster-specific manner. This approach allows the model to embed channels with different correlation profiles in separate latent subspaces, mitigating the risk of entangling unrelated features. 

Let \( d \) denote the desired final embedding dimension. 
Since the number of channels assigned to each cluster is not uniform, we assign a different embedding dimension \( d_i \) for each cluster \( i \in \{1, \dots, M\} \), proportional to the number of channels $C_i$ in that cluster, using the following formula,
\begin{align}
d_i = 
\begin{cases}
\left\lfloor \dfrac{C_i}{C}  d \right\rfloor, & \text{for } i = 1, \dots, M - 1 \\
d - \sum_{j=1}^{M{-}1} d_j, & \text{for } i = M,
\end{cases}
\end{align}
which ensures \(\sum_{i=1}^{M} d_i = d\). Additionally, all $d_i$ are lower-bounded by one due to either $d>C$ as in all experiments with benchmark datasets in Sec. \ref{experiments} or a manual constraint. Each channel cluster is processed through its respective embedding layer of dimension \( d_i \). These are subsequently concatenated to form the final unified \(d\)-dimensional embedding to feed into the next module of the model. 

The use of multiple embedders not only enhances the model’s capacity to learn localized representations within each cluster, it also reduces the number of trainable parameters in the embedding layer. In a conventional single linear embedding layer, the number of trainable parameters (weights) is \( C \times d \). In contrast, our approach partitions the channels into \( M \) clusters, and each cluster is embedded independently using a reduced embedding dimension. Assuming uniform cluster sizes, each cluster contains \( O\left(\frac{C}{M}\right) \) channels and is assigned an embedding dimension of \( O\left(\frac{d}{M}\right) \). Therefore, the number of trainable weights in each embedding layer is \(O\left(\frac{C \cdot d}{M^2}\right)\). Since we employ \( M \) embedding layers, the total number of trainable parameters becomes \(O\left(\frac{C \cdot d}{M}\right)\). 
Thus, the proposed method substantially reduces, by a factor of $M$, the number of trainable parameters in the embedding layer when \( M > 1 \).

\subsection{Causal Mixer}
\label{causal_mixer}



While, in existing MLP-Mixer models for time series analysis, each data point interacts with all others within the look-back window~\cite{wpmixer, tsmixer, timemixer}, 
we only let a data point at time $t$ 
interact with points from $\le t$ to maintain causality, mimicking the real-world systems. 
We use two mixing modules that
operate along the temporal dimension $L$ 
and the embedding dimension $d$ of the data space $\mathbb{R}^{L\times d}$. 

Note that the causality enforced in our model is purely temporal. Our objective is not to infer inter-channel causal relationships like Granger causality. We let the embedding mixer to learn 
useful patterns for $C$ channels.

\subsubsection{Temporal Mixer Module}
\label{causal_temporal_mixer_module}
This module consists of two causal linear layers connected by GELU activation. The input to the temporal mixer module is denoted by \(\vX_c \in \mathbb{R}^{d \times L}\), where \(L\) is the look-back window length and \(d\) is the embedding dimension. To introduce causality in the linear layer, we perform masking on the weight \(\vTheta_c \in \mathbb{R}^{L \times L}\) by an upper-triangular mask \(\vGamma \in \mathbb{R}^{L \times L}\) given by
{\small
\begin{equation}
    \begin{aligned}
        \vGamma &=
            \begin{bmatrix}
            \gamma_{1,1} & \gamma_{1,2} & \cdots & \gamma_{1,L} \\
            0          & \gamma_{2,2}  & \cdots & \gamma_{2,L} \\
            \vdots     & \vdots       & \ddots & \vdots      \\
            0          & 0            & \cdots & \gamma_{L,L}
            \end{bmatrix},
        \quad
        \gamma_{i,j} =
            \begin{cases}
            \frac{1}{j}, & \text{if } i \leq j \\
            0,           & \text{otherwise}
            \end{cases}
    \end{aligned}
\end{equation}
}
to get the final masked weight \(\vTheta_u \in \mathbb{R}^{L \times L}\):
\begin{align}
    \vTheta_u = \vGamma \odot \vTheta_c,
\end{align}
where \(\odot\) refers to element-wise multiplication.
Here, $1/j$ normalization accounts for the increasing number of aggregated past time steps at later positions, improving optimization stability during training, described in~\cref{ablation_1_over_j_mask} in the Appendix.
The operation in the first \emph{causal linear layer} can be written as,
\begin{align}
    \vx^{(k)}_{c1} = \vx^{(k)}_c \vTheta_u + \vb; \quad k = 1,\ldots,d
\end{align}
where $\vx^{(k)}_{c1} \in \mathbb{R}^{1 \times L}$ is the $k^{th}$ row vector of the output $\vX_{c1}$, $\vx^{(k)}_{c} \in \mathbb{R}^{1 \times L}$ is the $k^{th}$ row vector of the input $\vX_{c}$, and $\vb \in \mathbb{R}^{1 \times L}$ is the learnable bias. To enforce temporal causality, the causal linear layer restricts each output neuron \(j\) to depend only on input neurons \(i \leq j\), effectively preventing information flow from future neurons. We have another causal linear layer in the temporal mixer module, with operations similar to the first one. Figure~\ref{fig:neural_temporal_mixer} in Appendix provides a schematic illustration of the temporal mixer.


\subsubsection{Embedding Mixer}
\label{embedding_mixer}
The embedding mixer employs two linear layers with a GELU activation in between. It projects the input from \(d\) to \((d \cdot d_f)\) dimensions and then back to \(d\), following the design in~\cite{wpmixer}.

\subsection{Anomaly Detection Method}
\label{anomaly_detction_method}
Anomalies in time series data often exhibit sequential dependencies, where the likelihood of a point being anomalous increases if preceding points are anomalous. Leveraging this temporal association, we propose an anomaly detection method that considers not only the anomaly evidence of the current data point, but also that of preceding points. Our approach accumulates anomaly evidence over time, and when the accumulated evidence exceeds a predefined threshold, the corresponding instance is identified as anomalous. Furthermore, by analyzing the rise and fall of the anomaly evidence, our method effectively delineates the full extent of the anomaly sequence.

Let the original training and test datasets be denoted by \( \mathcal{X}_N \) and \(\mathcal{X}_T \), with sizes \( N \) and \( T \), respectively. We partition the training dataset \( \mathcal{X}_N \) into two subsets: a training subset \( \mathcal{X}_{N_1} \) and a validation subset \( \mathcal{X}_{N_2} \), such that \( N_1 + N_2 = N \). Our reconstruction model is trained on \( \mathcal{X}_{N_1} \) to minimize the reconstruction loss. After training, we compute the validation reconstruction loss for each sample in \( \mathcal{X}_{N_2} \). The reconstruction losses are then sorted in ascending order to form a sorted reconstruction loss series denoted by \( \mathcal{E} \in \mathbb{R}^{N_2} \). Similarly, the reconstruction loss series for the test dataset is denoted by \( \mathcal{G} = \{g_t\}_{t=1}^{T} \). 

To evaluate the anomaly likelihood of a test point \( g_t\), we calculate its empirical p-value (i.e., percentage of greater values) in \( \mathcal{E} \) as 
{
\small
$$
p_t = \frac{\#\{e\in\mathcal{E}:e\ge g_t\}}{\#\mathcal{E}},
$$
}
where $\#$ denotes the number of elements in a set. 
The anomaly evidence \( \beta_t \) is then computed using a statistical significance parameter \( \alpha \in [0, 1) \) as:
{
\small
\begin{align}
    \label{alpha}
    \beta_t = \log\left( \frac{\alpha}{p_t+\epsilon} \right).
\end{align}
}
where \(\epsilon\) is a small value that avoids division by zero. 
We then accumulate the anomaly evidence \( \beta_t \) over time using the following formula to get the accumulated evidence series,
{
\small
\begin{align}
\label{eq:anom_score}
    s_t = 
    \begin{cases}
        \max(s_{t-1} + \beta_t, 0), & \text{if } \neg \left( \bigwedge_{i=1}^{\delta} \left( \beta_{t-i} < 0 \right) \right) \\
        0, & \text{otherwise.}
    \end{cases}
\end{align}
}
Here, $s_t$ is initialized with $s_0 = 0$. The symbols \(\bigwedge\) and \(\neg\) are the logical AND and NOT, respectively. \( \delta \) is a predefined window length that controls the reset condition: bring $s_t$ to zero after observing $\delta$ consecutive negative anomaly evidences, potentially indicating the end of anomalous event. An instance is flagged as anomalous if \( s_t \) exceeds a predefined threshold \(h\). 
Let $t_a$ and $t_b$ denote the temporal index of the starting and endpoint of an initially detected anomaly segment. We further update $t_a$ and $t_b$ employing backward search, 
 \begin{flalign}
 & t_a^* = \arg\max\{ t \leq t_a \mid s_t = 0 \} \label{eq:anomaly_onset} \\
 & t_b^* = \arg\max\{ t_a \leq t \leq t_b \mid \beta_t > 0 \} \label{eq:anomaly_offset} 
 \end{flalign}
Here, $t_a^*$ and $t_b^*$ refer to the updated temporal index for the corresponding $t_a$ and $t_b$, respectively. The updated onset time \(t_a^*\) denotes the point at which the accumulated evidence begins to rise from zero before the alarm triggered by $s_t>h$, while the updated offset time \(t_b^*\) represents the point at which the accumulated evidence starts to decline near the end of the anomaly segment. A statistical interpretation of our anomaly detection method is provided in Appendix \ref{appendix_stat_interpretation_anom_method}, while the overall scoring process is illustrated in~\cref{fig:appendix_steps_anom_method} in the Appendix.


\begin{table}[!htb]
\footnotesize
\centering
\setlength{\tabcolsep}{4pt}
\caption{The statistics of the six publicly available datasets.}

\label{tab:dataset}
\begin{tabular}{@{}cccccc@{}}
Data. & Entities & Channels & Train & Test & Anom. Rate \\  \toprule
SWaT & 1 & 51 & 495000 & 449919 & 12.14\% \\
WADI & 1 & 123 & 1209601 & 172801 & 5.75\% \\
PSM & 1 & 25 & 132481 & 87841 & 27.76\% \\
SMD & 28 & 38 & 708405 & 708420 & 4.16\% \\
MSL & 27 & 55 & 58317 & 73729 & 10.48\% \\
SMAP & 55 & 25 & 140825 & 444035 & 12.83\% \\ \bottomrule
\end{tabular}
\vspace{-3mm}
\end{table}

\section{Experiments}
\label{experiments}
\paragraph{Datasets:} We evaluate the effectiveness of our model using six publicly available datasets. These datasets are: SWaT (Secure Water Treatment)~\cite{swat}, WADI (Water Distribution)~\cite{wadi}, PSM (Pooled Server Metrics)~\cite{psm}, SMD (Server Machine Dataset)~\cite{smd}, SMAP (Soil Moisture Active Passive)~\cite{smap}, and MSL (Mars Science Laboratory)~\cite{smap_msl}. The statistics of the datasets are shown in Table~\ref{tab:dataset}. The details of the datasets are discussed in~\cref{appendix_datasets} in the Appendix.
\paragraph{Baselines:} We compare our model against \camready{23 baselines}, including autoencoder-based, recurrent-based, generative, graph-based, transformer-based, MLP-mixer-based, and classical outlier detection approaches, reporting the best F1 scores in~\cref{tab:main_results}.\par
For multi-entity datasets, prior works adopt different evaluation protocols: (1) training a single model across all entities, (2) training separate models per entity and averaging entity-wise F1 scores, or (3) training per-entity models and aggregating predictions before computing F1. To ensure a fair and comprehensive comparison, we evaluate our model across all three protocols and report the results for each.\par
We additionally report SensitiveHUE under two settings: Offline and Online. The offline variant follows the original implementation, which uses robust normalization based on statistics computed over the entire test set, making it unsuitable for online inference. For fair comparison with our method, which performs online inference, we remove this post-hoc normalization to obtain the online SensitiveHUE variant. 
Detailed descriptions of all baselines and result sources are provided in Appendix \ref{appendix_baselines}.

\paragraph{Evaluation Metrics} 
\label{evaluation_metrics}
We use the best-F1 score as the primary evaluation metric, following prior works~\cite{npsr, sensitivehue}. The best-F1 score is obtained through a threshold-sweeping approach commonly adopted in the literature. Although some studies employ the point-adjusted F1 score~\cite{memto, huang2025graph}, we do not use it, as it disproportionately rewards detection performance for long-duration anomalies~\cite{doshi2022reward, npsr, kim2022towards, uae}. \camready{We additionally report PR\_AUC, which is particularly informative for imbalanced anomaly detection tasks~\cite{saito2015precision}}. Results on additional metrics, including VUS\_PR and VUS\_ROC~\cite{paparrizos2022volume}, are provided in the Appendix \ref{Additional_results}.

\begin{table}[!htb]
\footnotesize
\centering
\setlength{\tabcolsep}{2.5pt}
\caption{Performance comparison using the best-F1 score. For multi-entity datasets, marked with $^*$ (MSL, SMD, SMAP), methods annotated with a superscript $^{*i}$ are evaluated under protocol-$i$. The definitions of the evaluation protocols are provided in~\cref{experiments} (Baselines). Best (Bold) and second-best (Underlined) results are selected within each protocol for multi-entity datasets. Best-F1 score is defined in Appendix \ref{definition_f1}.}

\label{tab:main_results}
\begin{tabular}{@{}rcccccc@{}}
\toprule
            & WADI & PSM & SWAT & MSL$^*$ & SMD$^*$ & SMAP$^*$ \\ \midrule
$\text{Anom. Trans.}^{*1}$ & 0.108 & 0.434 & 0.220 & 0.191 & 0.080 & 0.227 \\
$\text{TimesNet}^{*1}$    & 0.322 & 0.436 & 0.260 & 0.261 & \underline{0.264} & 0.250 \\
$\text{xLSTMAD}^{*1}$     & 0.568 & 0.651 & 0.823 & 0.265 & 0.245 & 0.279 \\
$\text{CATCH}^{*1}$       & -     & 0.116 & 0.087 & 0.143 & 0.237 & 0.061 \\
$\text{D3R}^{*1}$         & 0.129 & 0.442 & 0.459 & 0.240 & - & 0.227 \\
$\text{PatchAD}^{*1}$     & 0.528 & 0.493 & 0.777 & 0.249 & - & 0.247 \\
$\text{SimAD}^{*1}$       & 0.640 & 0.521 & 0.820 & \underline{0.300} & - & \underline{0.294} \\ 
$\text{TimesNet}^{*2}$    & -     & -     & -     & 0.369 & 0.477 & 0.354 \\
$\text{xLSTMAD}^{*2}$     & -     & -     & -     & 0.441 & 0.521 & 0.451 \\
$\text{DAGMM}^{*2}$       & 0.121 & 0.483 & 0.750 & 0.199 & 0.238 & 0.333 \\
$\text{LSTM-VAE}^{*2}$    & 0.227 & 0.455 & 0.776 & 0.212 & 0.435 & 0.235 \\
$\text{MSCRED}^{*2}$      & 0.046 & 0.556 & 0.757 & 0.250 & 0.382 & 0.170 \\
$\text{OmniAnom.}^{*2}$   & 0.223 & 0.452 & 0.782 & 0.207 & 0.474 & 0.227 \\
$\text{MAD-GAN}^{*2}$     & 0.370 & 0.471 & 0.770 & 0.267 & 0.220 & 0.175 \\
$\text{MTAD-GAT}^{*2}$    & 0.437 & 0.571 & 0.784 & 0.275 & 0.400 & 0.296 \\
$\text{USAD}^{*2}$        & 0.233 & 0.479 & 0.792 & 0.211 & 0.426 & 0.228 \\
$\text{UAE}^{*2}$         & 0.354 & 0.427 & 0.453 & 0.451 & 0.435 & 0.390 \\
$\text{GDN}^{*2}$         & 0.570 & 0.552 & 0.810 & 0.217 & 0.529 & 0.252 \\
$\text{TranAD}^{*2}$      & 0.415 & 0.649 & 0.669 & 0.251 & 0.310 & 0.247 \\
$\text{NPSR}^{*2}$        & 0.642 & 0.648 & 0.839 & \underline{0.551} & 0.535 & \underline{0.505} \\
$\text{SAT}^{*2}$         & 0.635 & 0.653 & 0.842 & 0.503 & 0.506 & 0.452 \\
$\text{OracleAD}^{*2}$    & - & \underline{0.659} & 0.765 & - & 0.430 & - \\
$\text{PCA-Error}^{*2}$   & - & - & 0.833 & - & \underline{0.572} & - \\ 
$\text{Sen.H.(Off)}^{*3}$ & \underline{0.700} & 0.517 & \textbf{0.911} & \underline{0.452} & \underline{0.398} & 0.415 \\
$\text{Sen.H.(On)}^{*3}$  & 0.678 & 0.507 & 0.874 & 0.418 & 0.346 & \underline{0.523} \\ 
$\text{CAROTS}$           & 0.143 & 0.603 & 0.791 & -     & -     & - \\ 
$\text{CAROTS+}$           & 0.391 & 0.534 & 0.789 & -     & -     & - \\ \midrule
$\text{CCM-TAD}^{*1}$     & \textbf{0.761} & \textbf{0.716} & \underline{0.883} & \textbf{0.370} & \textbf{0.300} & \textbf{0.389} \\
$\hspace{1.5cm}^{*2}$     & \textbf{-} & \textbf{-} & \textbf{-} & \textbf{0.613} & \textbf{0.612} & \textbf{0.529} \\
$\hspace{1.5cm}^{*3}$     & \textbf{-} & \textbf{-} & \textbf{-} & \textbf{0.638} & \textbf{0.602} & \textbf{0.551} \\ \bottomrule
\end{tabular}
\vspace{-3mm}
\end{table}

\subsection{Results}
\label{results}
Table~\ref{tab:main_results} presents the best-F1 scores across six publicly available datasets. Our model consistently outperforms existing methods on nearly all datasets. It improves the best-F1 scores by 8.71\% and 8.65\% for WADI and PSM, respectively. For the SWAT dataset, although SensitiveHUE (Offline) slightly outperforms our model, our method surpasses SensitiveHUE (Online), which is a more appropriate baseline given that our model detects anomalies online. We also outperform all baselines across multi-entity datasets under each evaluation protocol. For MSL, the improvements are $+23.3\%(0.300\to0.370)$, $+11.3\%(0.551\to0.613)$, and $+41.2\%(0.452\to0.638)$ across the three protocols, respectively. For SMAP, the improvements are $+32.3\%(0.294\to0.389)$, $+4.8\%(0.505\to0.529)$, and $+5.4\%(0.523\to0.551)$ across the three protocols, respectively. Furthermore, the improvements in the best-F1 score for the SMD dataset are also consistent. We observe that training a model for each entity yields better results than training a single model for all entities, as entity-entity variations are considerable.\par
\camready{In addition, Table~\ref{tab:pr_auc} presents the PR\_AUC results across the datasets. Our model consistently outperforms the baselines on almost all datasets by a considerable margin. Although CCM-TAD does not achieve the best PR\_AUC on the PSM dataset, its performance remains competitive.}

\begin{table*}[]
\footnotesize
\centering
\caption{\camready{Performance comparison using PR AUC. Multi-entity datasets annotated with superscript$^{*i}$
correspond to results under evaluation protocol-$i$. Best and second-best results are shown in bold and underline, respectively, within each protocol for multi-entity datasets. Some methods provide results only on a subset of datasets; therefore, some entries are unavailable.}}
\label{tab:pr_auc}
\begin{tabular}{@{}cccc|ccc|ccc@{}}
\toprule
 & WADI & PSM & SWAT & MSL$^{*1}$ & SMD$^{*1}$ & SMAP$^{*1}$ & MSL$^{*2}$ & SMD$^{*2}$ & SMAP$^{*2}$ \\ \midrule
TimesNet & 0.250 & 0.368 & 0.155 & \underline{0.183} & \underline{0.183} & 0.113 & 0.247 & 0.431 & 0.208 \\
xLSTMAD & 0.550 & \textbf{0.598} & 0.805 & 0.178 & 0.136 & \underline{0.132} & \underline{0.312} & \underline{0.457} & \underline{0.353} \\
CATCH & - & 0.434 & 0.166 & 0.167 & 0.172 & 0.131 & - & - & - \\
CAROTS & 0.056 & \underline{0.595} & 0.764 & - & - & - & - & - & - \\
CAROTS+ & 0.260 & 0.535 & 0.760 & - & - & - & - & - & - \\
Sen.H.(On) & \underline{0.612} & 0.460 & \underline{0.852} & - & - & - & 0.283 & 0.407 & 0.346 \\ \midrule
CCM-TAD & \textbf{0.720} & 0.558 & \textbf{0.911} & \textbf{0.283} & \textbf{0.311} & \textbf{0.336} & \textbf{0.610} & \textbf{0.587} & \textbf{0.552} \\ \bottomrule
\end{tabular}
\end{table*}

\begin{table*}[!htb]
\small
\centering
\setlength{\tabcolsep}{4pt}
\caption{Contribution from the proposed sequential anomaly detection method (Sec. \ref{anomaly_detction_method}). ``Seq. w/o BC'' denotes our sequential anomaly scoring method without boundary correction (Eqs. ~\ref{eq:anomaly_onset} and ~\ref{eq:anomaly_offset}), while ``Seq. w/ BC'' corresponds to the method with boundary correction.}
\label{tab:ablation_anomaly_detection_frame}
\begin{tabular}{@{}r|cc|cc|cc|cc|cc@{}}
\toprule
 & \multicolumn{2}{c|}{SWAT} & \multicolumn{2}{c|}{PSM} & \multicolumn{2}{c|}{WADI} & \multicolumn{2}{c|}{MSL} & \multicolumn{2}{c}{SMD} \\ \midrule
 & F1 & PR\_AUC
 & F1 & PR\_AUC
 & F1 & PR\_AUC
 & F1 & PR\_AUC
 & F1 & PR\_AUC \\ \midrule
Point-based & \underline{0.882} & 0.869 & 0.657 & \underline{0.536} & \underline{0.739} & 0.660 & 0.557 & 0.339 & 0.562 & 0.509 \\
Seq. w/o BC & 0.879 & \underline{0.888} & \underline{0.659} & 0.525 & \underline{0.739} & \underline{0.678} & \underline{0.626} & \underline{0.498} & \underline{0.595} & \underline{0.530} \\
Seq. w/ BC & \textbf{0.883} & \textbf{0.911} & \textbf{0.716} & \textbf{0.558} & \textbf{0.761} & \textbf{0.720} & \textbf{0.638} & \textbf{0.610} & \textbf{0.602} & \textbf{0.587} \\ \bottomrule
\end{tabular}
\vspace{-3mm}
\end{table*}

\begin{table}[!htb]
\small
\centering
\setlength{\tabcolsep}{4pt}
\caption{Ablation study evaluating the impact of channel clustering, temporal mixer, and embedding mixer modules. The symbols \(\times\) and \(\checkmark\) denote the presence and absence of each module. The symbols \(\diamondsuit\) and \(\spadesuit\) indicate non-causal and causal temporal mixers, respectively. Reported values are the best F1 scores obtained. Case 10 represents the complete model configuration.}
\label{tab:ablation_modules_new}
\begin{tabular}{@{}lccccc@{}} 
\toprule
Case 
& Clustering 
& Temporal Mx 
& Embed. Mx 
& SWAT 
& PSM \\ 
\midrule
1  & $\times$ & $\times$              & \checkmark & 0.866 & 0.670 \\
2  & $\times$ & \(\diamondsuit\)      & $\times$   & 0.873 & 0.687 \\
3  & $\times$ & \(\diamondsuit\)      & \checkmark & 0.871 & 0.673 \\
4  & $\times$ & \(\spadesuit\)        & $\times$   & 0.861 & 0.653 \\
5  & $\times$ & \(\spadesuit\)        & \checkmark & 0.865 & 0.703 \\
6  & \checkmark & $\times$            & \checkmark & 0.874 & 0.657 \\
7  & \checkmark & \(\diamondsuit\)    & $\times$   & 0.872 & 0.661 \\
8  & \checkmark & \(\diamondsuit\)    & \checkmark & 0.871 & 0.652 \\
9  & \checkmark & \(\spadesuit\)      & $\times$   & 0.875 & 0.681 \\
10 & \checkmark & \(\spadesuit\)      & \checkmark & \textbf{0.883} & \textbf{0.716} \\
\bottomrule
\end{tabular}
\end{table}

\begin{table}[!htb]
\centering
\small
\setlength{\tabcolsep}{3pt}
\caption{Nearest-neighbor separation (NN-sep) across datasets. 
Imp(\%) denotes the relative improvement of the proposed clustering approach compared to the model without clustering.}

\label{tab:nn_sep}
\begin{tabular}{ccccc}
\toprule
Dataset & 
w/o Clustering & w/ Clustering & Imp.(\%)\\
\midrule
SWAT & 
65.63 & \textbf{91.30}  & 39.11\\
WADI & 
12.01 & \textbf{12.47}  & 3.83 \\
PSM  & 
66.74 & \textbf{66.78}  & 0.06 \\
\bottomrule
\end{tabular}
\end{table}

\begin{table}[!htb]
\small
\centering
\caption{Best F1 scores for various clustering variants. The number of clusters \(M\) is optimized for each variant. Variants (i), (ii), (iii) are explained in Section~\ref{ablation_clustering_matin_paper} and Appendix \ref{ablation_additional_channel_clustering_approaches}.} 
\label{tab:different_clustering_methods}
\begin{tabular}{@{}ccccc@{}}
\toprule
 Variant & Proposed & (i) & (ii) & (iii) \\ \midrule
Dataset & \(M (F1)\) & \(M (F1)\) & \(M (F1)\) & \(M (F1)\) \\ \midrule
PSM & 4 (\textbf{0.716}) & 3 (0.688) & 5 (0.674) & 6 (0.673)  \\
SWAT & 6 (\textbf{0.883}) & 4 (0.873) & 4 (0.879) & 6 (0.881) \\ \bottomrule
\end{tabular}
\vspace{-3mm}
\end{table}

\subsection{Ablation Study}
\label{ablation_study}
In the ablation studies, hyperparameters are tuned for each experiment separately. Further details of hyperparameters are provided in Appendix~\ref{training_details}. Additional ablation studies are presented in Appendix \ref{Additional_Ablation_Studies}.

\subsubsection{Anomaly Detection Method}
\label{ablation_anomaly_detection}
\begin{figure}[!htb]
    \centering
    {\scriptsize
    \textcolor{blue}{—} Seq. Anom. Score \quad
    \textcolor{green!50!black}{—} Reconstruction loss \quad
    \fcolorbox{red}{red!20}{\phantom{\rule{3pt}{3pt}}} True Anom. Region
    }
    \includegraphics[width=0.95\columnwidth]{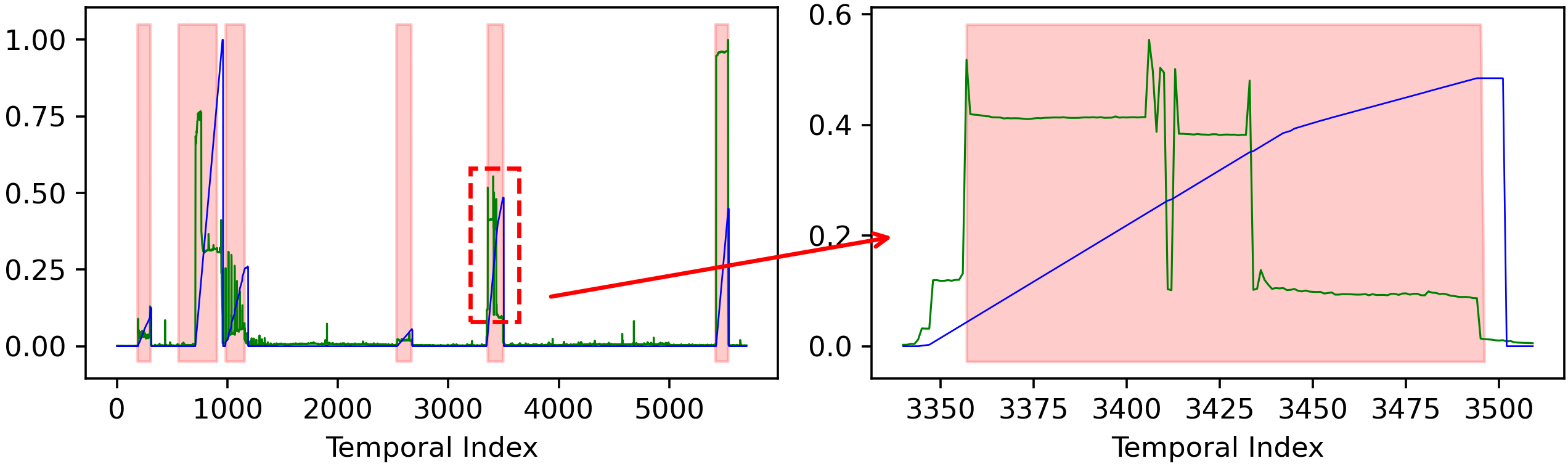}
    \caption{Visualization of point-based anomaly score (Recons. loss) and our proposed sequential anomaly score. The proposed sequential anomaly score produces smooth, temporally coherent responses across anomaly regions, thereby improving detection performance.}
    \label{fig:visualizatoin_anom_score}
    \vspace{-3mm}
\end{figure}

To evaluate the effectiveness of our proposed sequential anomaly detection method, we compare its performance against the point-based anomaly detection version of our method. In point-based anomaly detection, we employ the reconstruction loss (MSE) as the anomaly score and compare it to a threshold $h$ to detect anomalies. As shown in~\cref{tab:ablation_anomaly_detection_frame}, our method without boundary correction (Equations ~\ref{eq:anomaly_onset} and ~\ref{eq:anomaly_offset}) already outperforms the point-based detection across most datasets. The performance is further improved by boundary correction. One reason for this improvement is that, unlike the point-based approach, which may classify an anomalous instance as normal even when adjacent anomalies are correctly detected, our method accumulates anomaly evidence over time. As a result, the likelihood of missing anomalous instances within a contiguous sequence is substantially reduced, as illustrated in~\cref{fig:visualizatoin_anom_score}.


\subsubsection{Modular Architecture}
\label{ablation_modular}
In Table~\ref{tab:ablation_modules_new}, we present 10 cases, showing different configurations of our model, to demonstrate the impact of different modules on the model's performance.
It is observed that the channel clustering and the causal temporal mixer modules are complementary to each other. Case-10 corresponds to the full configuration of our model. By comparing Case-5 and Case-10, it is observed that clustering improves the best F1 score on both SWAT and PSM datasets significantly. We also evaluate the performance of our model with a regular temporal mixer (non-causal), our proposed causal temporal mixer, and without a temporal mixer. It is observed that the performance with the non-causal temporal mixer (Case-8) and without any temporal mixer (Case-6) have similar performances. We can conclude that the non-causal temporal mixer does not effectively capture the temporal information for anomaly detection. However, the proposed causal temporal mixer (Case-10) significantly improves F1 score, compared to Case-6 and Case-8. 

\subsubsection{Effect of Clustering on Representation Separability}
\label{ablation_NN_seperation}
We quantify how clustering affects the geometry of the learned representations using the nearest-neighbor separation ratio \(\text{NN-sep} = \frac{\mathbb{E}[d(a\!\rightarrow\! n)]}{\mathbb{E}[d(n\!\rightarrow\! n)]}\), where \(d(a\!\rightarrow\! n)\) is the distance from each anomalous sample to its nearest normal neighbor and \(d(n\!\rightarrow\! n)\) is the nearest-neighbor distance among normals. Here \emph{nearest} refers to Euclidean proximity in the representation space $\mathbb{R}^C$. Higher NN-sep values indicate a larger normal–anomaly margin. As shown in Table~\ref{tab:nn_sep}, clustering substantially enlarges this margin (91.30) on SWAT, compared to the 
embedding without clustering (65.63), indicating a +39\% improvement over without clustering. For WADI, this improvement is 3.83\%, while it is 0.06\% for PSM. These results demonstrate that cluster-aware embedding enhances the discrimination between normal and anomalous representations.

\subsubsection{Variants of Channel Clustering}
\label{ablation_clustering_matin_paper}
To assess the robustness of our spectral clustering approach, we compare it with three alternative methods: (i) spectral clustering based on cosine similarity between channels, (ii) spectral clustering using the absolute correlation matrix, and (iii) K-Means clustering applied directly to the correlation profiles. Table~\ref{tab:different_clustering_methods} shows the performance comparison of these variants with optimized number of clusters $M$. Detailed descriptions for these variants are provided in Appendix~\ref{ablation_additional_channel_clustering_approaches}. Our approach consistently outperforms these baselines, demonstrating the effectiveness of spectral clustering based on the full correlation profile.

\begin{figure}[!htb]
  \centering
  \subfloat[Non-causal\label{fig:grad_noncausal}]{\includegraphics[width=0.5\columnwidth]{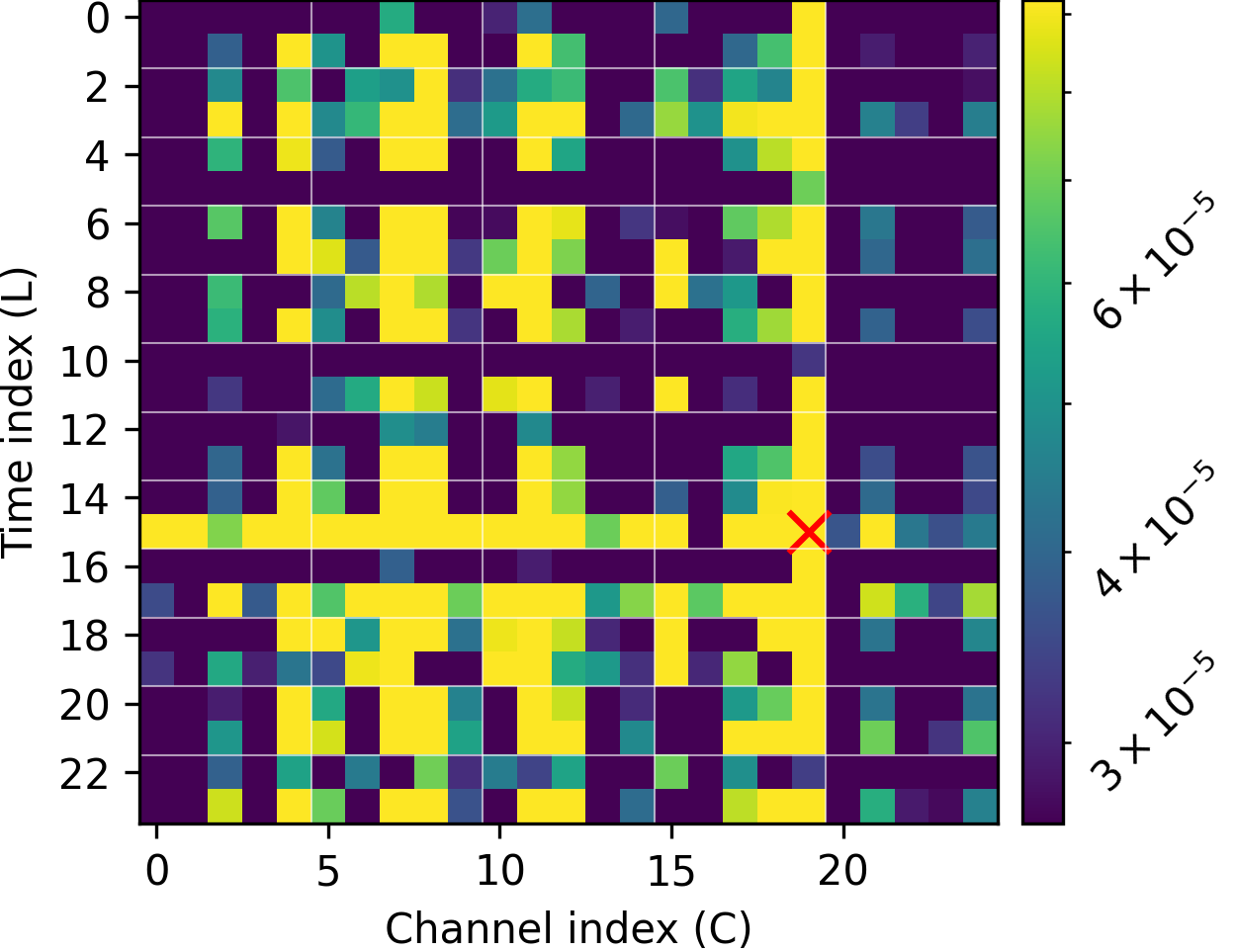}}\hfill
  \subfloat[Causal\label{fig:grad_causal}]{\includegraphics[width=0.5\columnwidth]{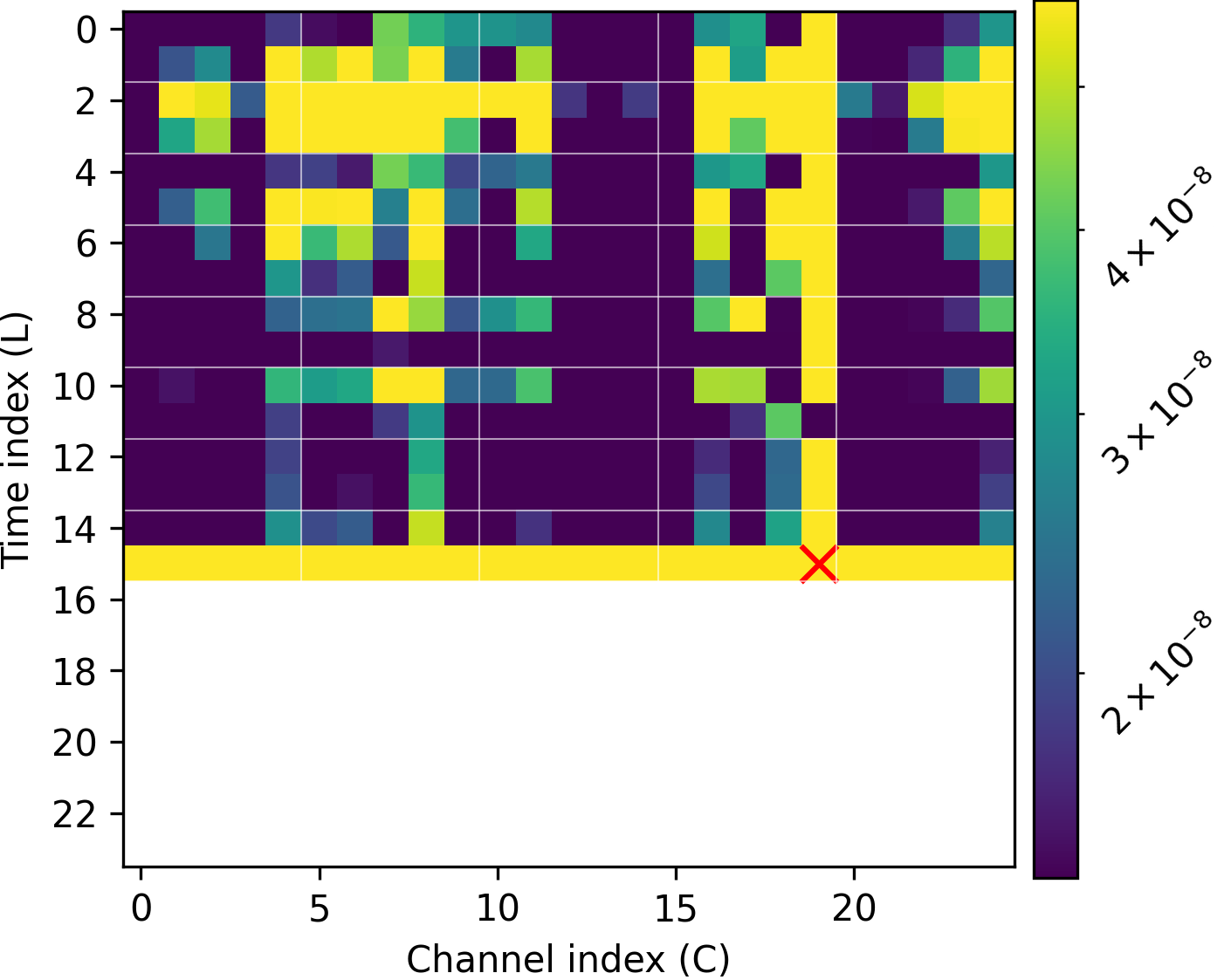}} 
  \caption{Gradient-based temporal dependency analysis for causal and non-causal temporal mixers. The heatmaps show the magnitude of the input gradients of the reconstruction loss evaluated at time index $t=15$ and channel $c=19$ with respect to all input time--channel pairs of $\mathbf{X}_L$. The red ``$\times$'' denotes the loss evaluation point. 
  }

  \label{fig:causal_grad}
  \vspace{-5mm}
\end{figure}

\subsubsection{Verifying Temporal Causality in the Model via Gradient Analysis}
\label{ablation_temporal_causality_in_the_model}

To confirm that the proposed model maintains temporal causality, we conduct an experiment using two variants of the model: one employing a causal linear layer in the temporal mixer and the other using a standard (non-causal) linear layer. We analyze the input-gradient dependencies of the reconstruction loss. For the input $\mathbf{X}_L \in \mathbb{R}^{L \times C}$, we evaluate the reconstruction loss at time index $t = 15$ and channel $c = 19$, and compute the absolute input gradients $\left| \partial \ell / \partial x^{j}_{\tau} \right|$ over all time--channel pairs $(\tau, j)$.

Figure~\ref{fig:causal_grad} compares the resulting gradient maps for the models with causal and non-causal temporal mixers. The model with the non-causal temporal mixer exhibits non-zero gradients for future time steps ($\tau > 15$), indicating that the reconstruction at time $t$ depends on future inputs. In contrast, the model with the causal temporal mixer completely suppresses gradients for all $\tau > 15$, confirming that the loss at time $t$ is influenced only by present and past observations.

\section{Conclusion}

In this work, we present CCM-TAD, a unified time-series anomaly-detection framework. Our approach combines multiple complementary novelties that jointly address fundamental limitations of existing methods. First, we introduce temporal causality into the MLP-Mixer architecture, using a principled causal masking design. Extensive ablations and gradient-based analyses confirm that causal temporal mixing significantly helps anomaly detection and completely eliminates future information leakage. We also propose a correlation-profile-based channel clustering strategy and incorporate it into a cluster-aware multi-embedding module. This effectively mitigates spurious inter-channel correlations in the normal data, improves the separability of normal–anomalous representations, and consistently outperforms alternative clustering approaches. We further propose a sequential anomaly scoring method with boundary refinement that accumulates statistically grounded anomaly evidence over time. The proposed scoring mechanism captures temporal continuity, refines anomaly boundaries, and enables online anomaly detection. Together, these contributions form a coherent and effective anomaly detection framework that is causal, robust, and suitable for real-time deployment. Extensive experiments on six public benchmarks demonstrate consistent improvements over the baselines across both single-entity and multi-entity datasets.

\section*{Impact Statement}
This paper presents work whose goal is to advance the field of Machine Learning, specifically online anomaly detection in time series. There are many potential societal consequences of our work; we do not feel that any must be specifically highlighted beyond encouraging responsible deployment with appropriate oversight in high-stakes monitoring settings.

\bibliography{egbib}
\bibliographystyle{icml2026}

\newpage
\appendix
\onecolumn
\appendix

\section{Statistical Interpretation of our Anomaly Detection Method}
\label{appendix_stat_interpretation_anom_method}

\begin{figure}[!htb]
  \centering
  \subfloat[Reconstruction loss\label{fig:step_1_anomaly_detection}]{\includegraphics[width=0.5\textwidth]{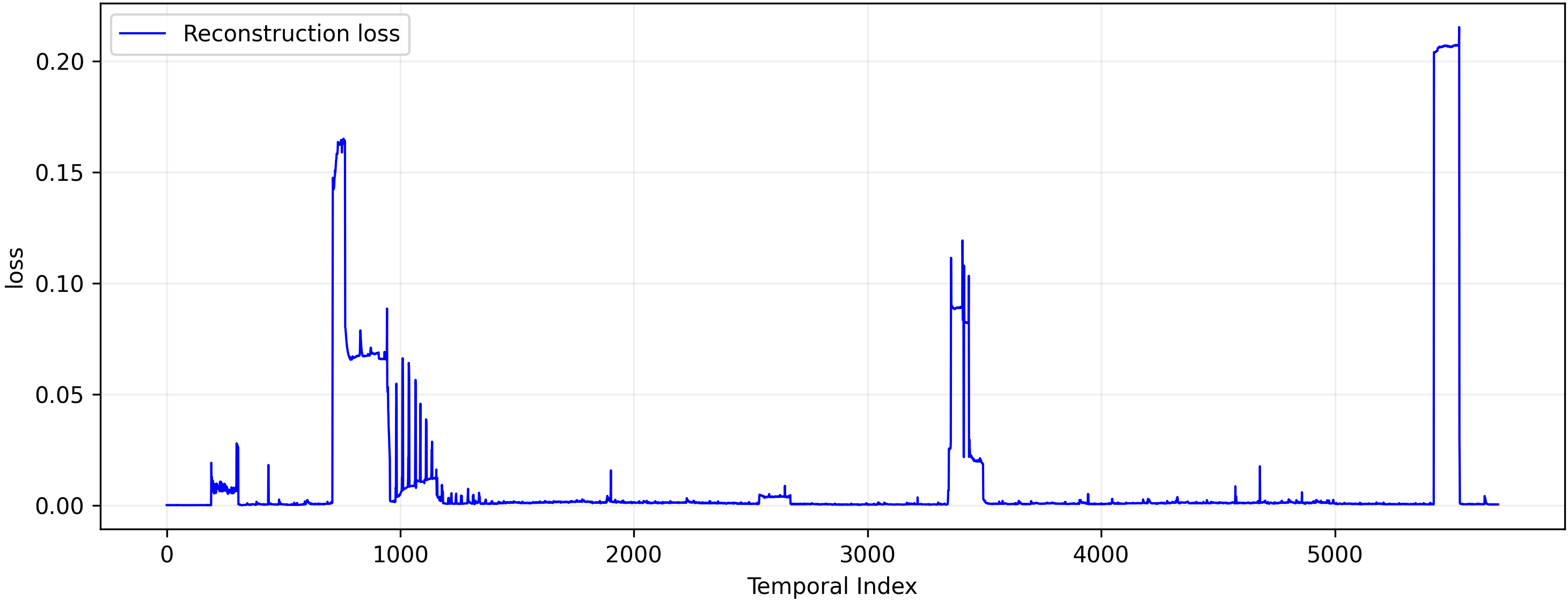}}\hfill
  \subfloat[$p$-value\label{fig:step_2_anomaly_detection}]{\includegraphics[width=0.5\textwidth]{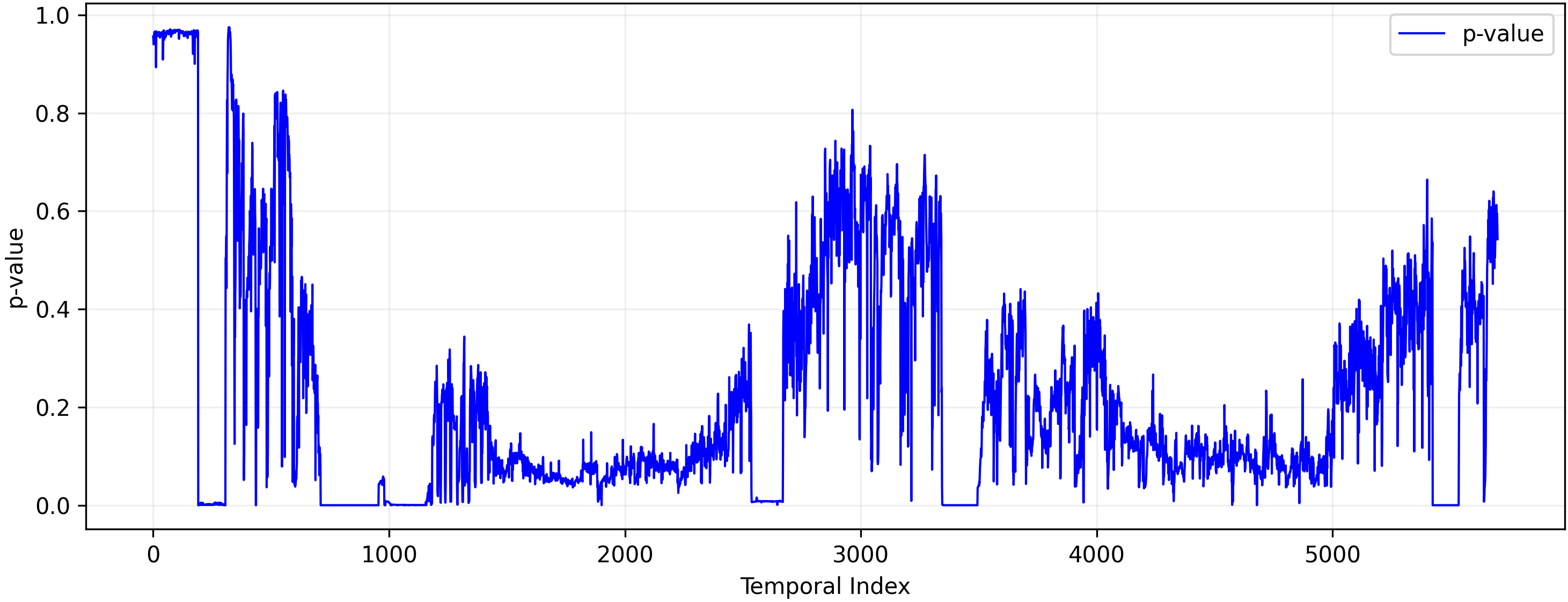}}\vfill
  \vspace{5pt}
  \subfloat[Anomaly evidence $\beta_t$\label{fig:step_3_anomaly_detection}]{\includegraphics[width=0.5\textwidth]{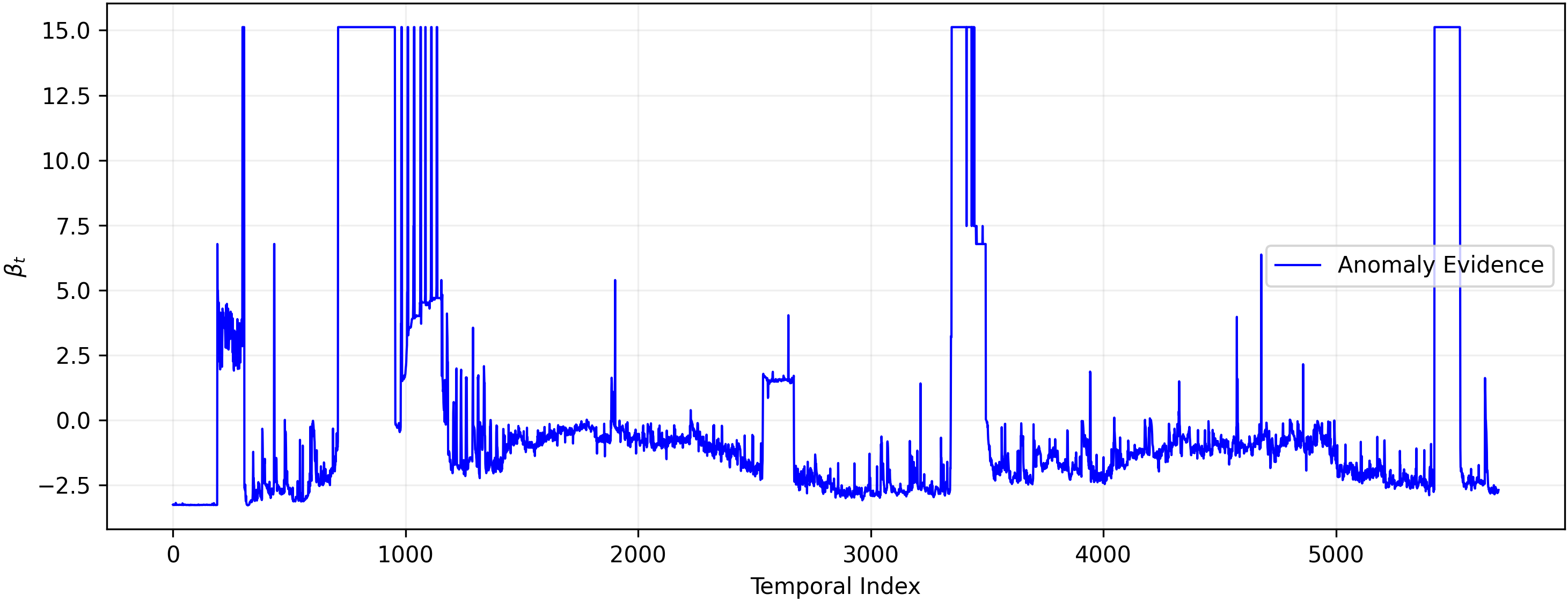}}\hfill
  \subfloat[Accumulated Anom. evidence $s_t$\label{fig:step_4_anomaly_detection}]{\includegraphics[width=0.5\textwidth]{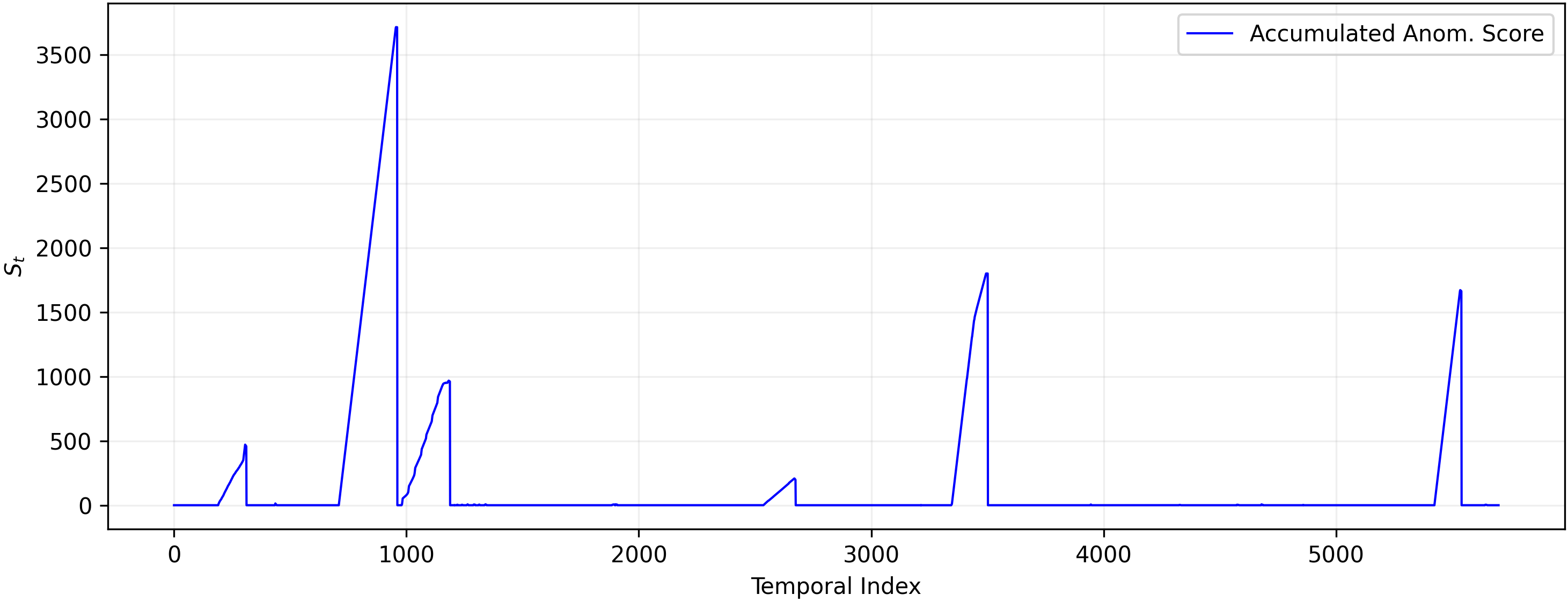}}\vfill
  \vspace{5pt}
  \subfloat[Final Prediction (GT: Ground Truth, Pred: Prediction)\label{fig:step_5_anomaly_detection}]{\includegraphics[width=0.5\textwidth]{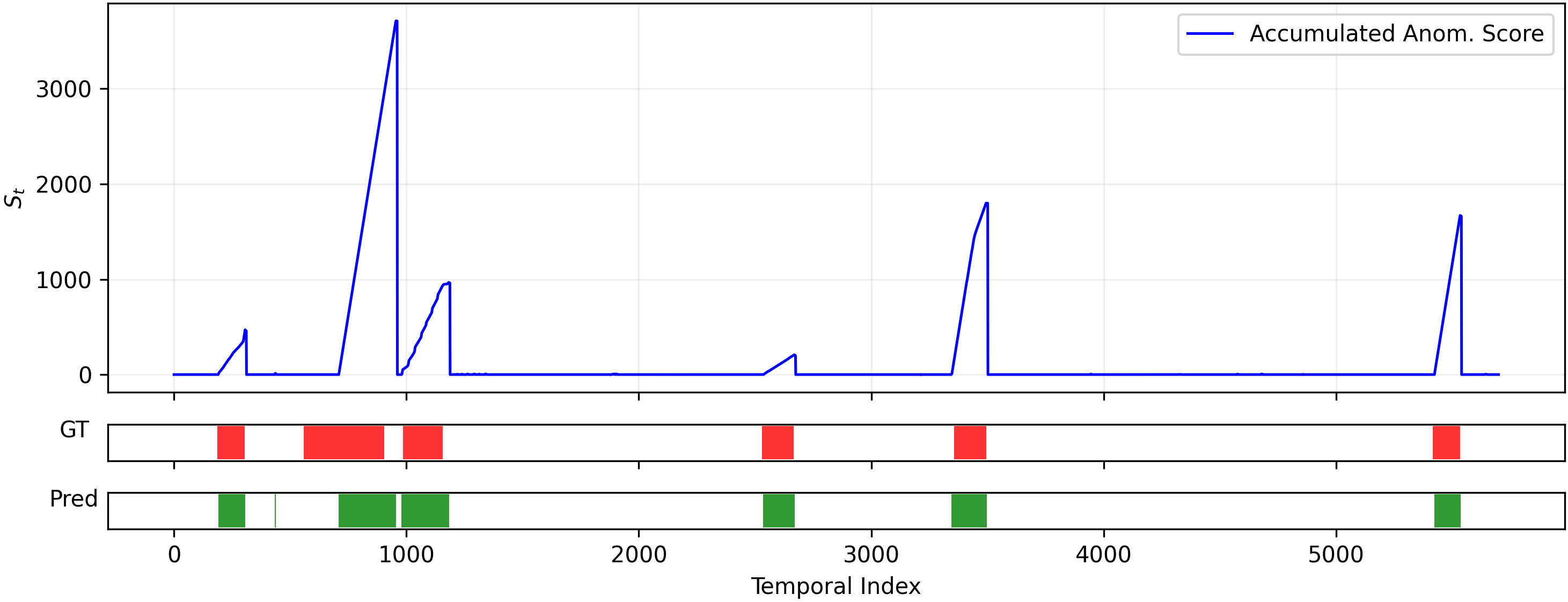}}
  \caption{Overview of the proposed anomaly detection scoring process. The reconstruction loss is converted to a p-value, which is then used to generate anomaly evidence $\beta_t$ and accumulated over time to produce the final anomaly score $s_t$ and predictions. Anomaly detection method is detailed in~\cref{anomaly_detction_method} in the main paper.}
  \label{fig:appendix_steps_anom_method}
\end{figure}

Here, we illustrate how our method operates in Fig. \ref{fig:appendix_steps_anom_method} and analyze how the anomaly score $s_t$ (Eq. \eqref{eq:anom_score}) is expected to behave statistically under normal and anomalous data. Since $s_t$ is a running sum of anomaly evidence $\beta_t=\log(\alpha/(p_t+\epsilon))$ over time, we analyze the probability distribution and expectation of $\beta_t$ under normal and anomalous data, i.e., $f_0(\beta)$ and $f_1(\beta)$, $\mathbb{E}_0[\beta]$ and $\mathbb{E}_1[\beta]$, respectively. 

\begin{figure}[h]
    \centering
    \includegraphics[width=0.49\linewidth]{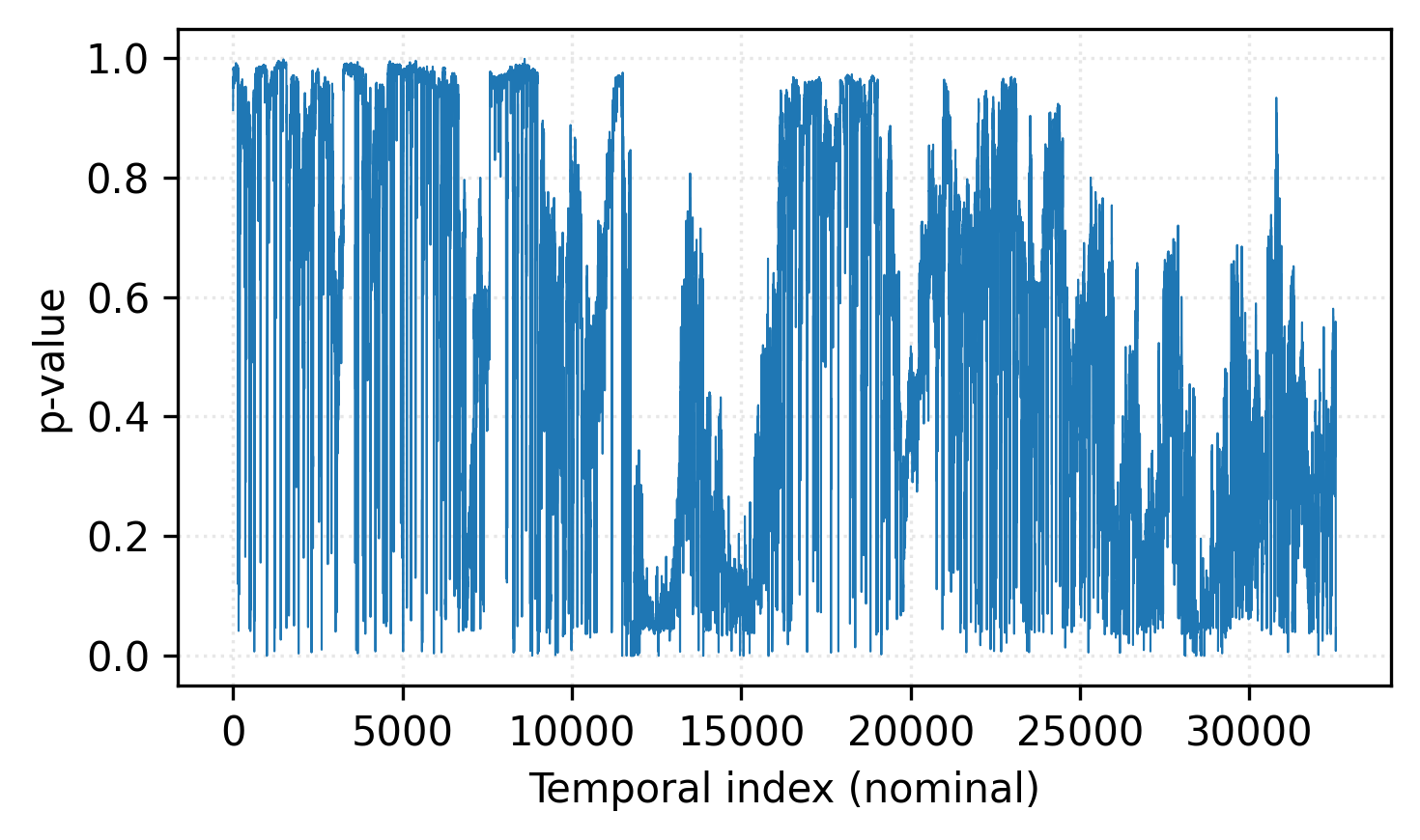}
    \includegraphics[width=0.49\linewidth]{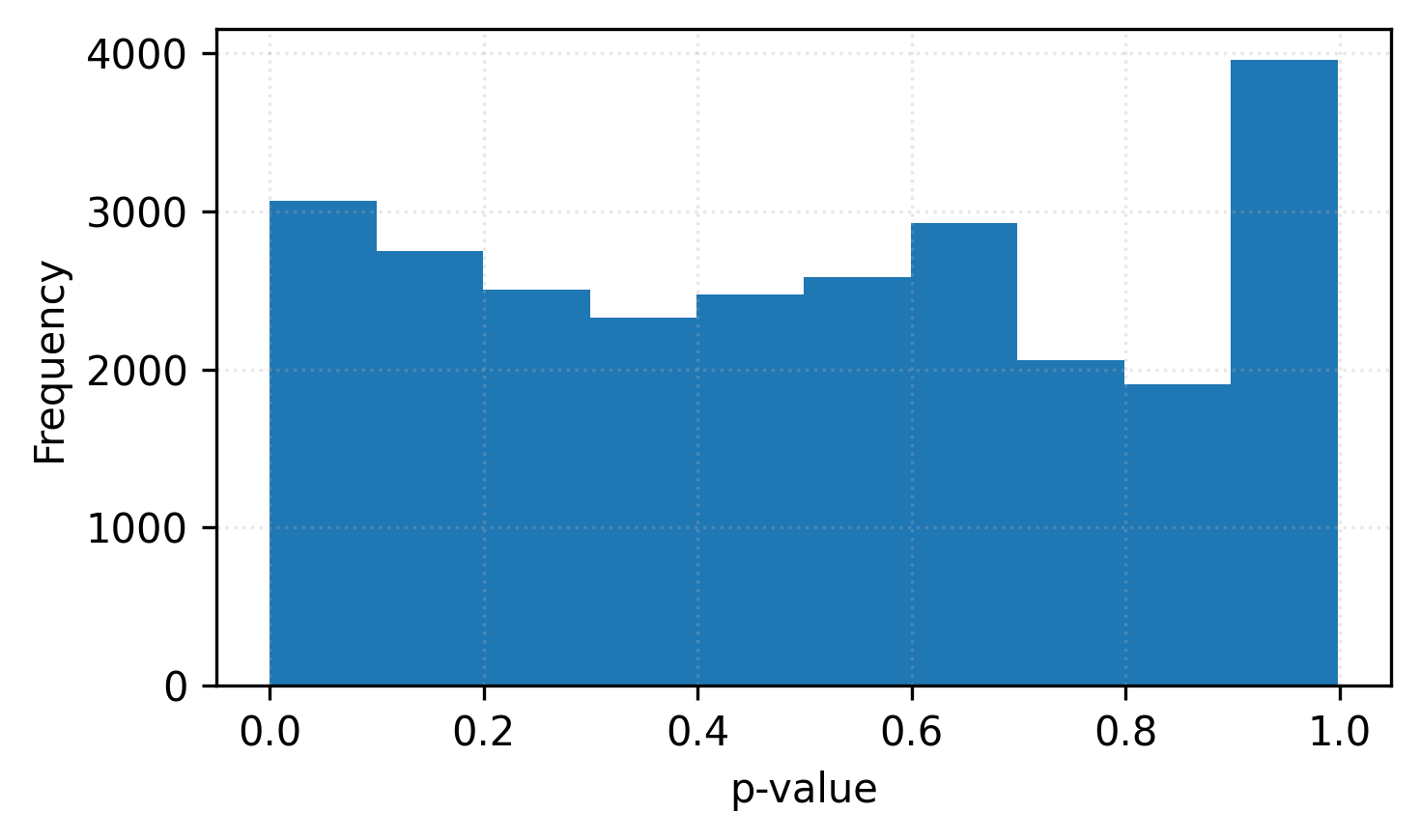}
    \caption{Temporal evolution of p-values on the normal WADI test data (left) and the corresponding empirical distribution shown as a histogram (right).}
    \label{fig:p-value-series}
\end{figure}

The empirical p-value $p_t$ converges to the actual p-value as the validation set size $N_2$ grows. The p-value of independent and identically distributed (iid) samples are known to be uniformly distributed in $[0,1]$ regardless of the continuous probability distribution of samples \cite{casella2024statistical}. Although the objective of reconstruction model is to capture all learnable pattern and leave only iid noise as reconstruction loss, in general, for time series data, reconstruction loss, and in turn its p-value, also constitute time series with temporal dependency, as shown in Fig. \ref{fig:p-value-series}. Despite the temporal correlation, the p-value of reconstruction loss approximately takes uniformly random values in $[0,1]$, as illustrated in the histogram in Fig. \ref{fig:p-value-series}, with possibly correlated durations around some value (e.g., initially values mostly close to $1$ in Fig. \ref{fig:p-value-series}). Note that the quality of this approximation depends on the dataset, as well as the performance of reconstruction model. 

Approximating the distribution of empirical p-value $p_t$ with $\mathcal{U}(0,1)$, we write the cumulative distribution function (cdf) of $\beta_t$ as 
\begin{align*}
    F_0(\beta) &= \mathbb{P}(\beta_t\le \beta) = \mathbb{P}\left( \log\frac{\alpha}{p_t+\epsilon} \le \beta \right) \\
    &= \mathbb{P}\left( p_t \ge \frac{\alpha}{e^\beta}-\epsilon \right) \\ 
    &\approx \begin{cases}
        1-\frac{\alpha}{e^\beta}+\epsilon, ~~\text{if}~~ \beta \in [\log \frac{\alpha}{1+\epsilon}, \log\frac{\alpha}{\epsilon}] \\
        0,~~\text{otherwise}.
    \end{cases}
\end{align*}
Taking the derivative of cdf we obtain the probability density function (pdf) of $\beta_t$ as follows:
\begin{align*}
    f_0(\beta)=\frac{\partial F_0(\beta)}{\partial \beta} \approx \begin{cases}
        \alpha e^{-\beta}, ~~\text{if}~~ \beta \in [\log \frac{\alpha}{1+\epsilon}, \log\frac{\alpha}{\epsilon}] \\
        0,~~\text{otherwise}.
    \end{cases}
\end{align*}
Thus, we can approximate the expectation under normal data as 
\begin{align*}
    \mathbb{E}_0[\beta] &= \int_{\log \frac{\alpha}{1+\epsilon}}^{ \log\frac{\alpha}{\epsilon}} \beta \alpha e^{-\beta} \text{d}\beta 
    = \left(\log \frac{\alpha}{1+\epsilon}+1\right)(1+\epsilon) \underbrace{- \left(\underbrace{\log \frac{\alpha}{\epsilon}}_{>0}+1\right)\epsilon}_{<0},
\end{align*}
which is $<0$ for $\frac{\alpha}{1+\epsilon}<e^{-1}$, i.e., $\alpha < \frac{1+\epsilon}{e}$. Hence, when data is normal and from the same distribution as training and validation, the anomaly evidence $\beta_t$ is expected to be negative for $\alpha<e^{-1}\approx 0.36788$ and in turn the running anomaly score $s_t$ is expected to hover around zero. On the other hand, when data becomes anomalous, reconstruction loss is expected to grow such that $p_t$ becomes smaller than $\alpha$, making $\beta_t$ positive and $s_t$ grow.

\section{Spectral Channel Clustering Procedure}
\label{app:cluster}
A pseudocode for the spectral channel clustering algorithm is given in Algorithm \ref{algo:channel_clustering}.

\begin{algorithm}[!hb]
\caption{\textsc{Proposed Spectral Clustering Based on Correlation Profile}}
\label{algo:channel_clustering}
\begin{algorithmic}[1]
\Function{SpectralClustering}{$\mathcal{X}_N \in \mathbb{R}^{N \times C}, M < C \in \mathbb{N}$}
    \State \textbf{Input:} Training dataset \(\mathcal{X}_N\), Number of clusters $M$
    \State \textbf{Output:} Cluster index for $C$ channels $k \in \{1, 2, \ldots, M\}^C$
    
    \State $\vPhi \gets \textsc{CorrCoeffMatrix}(\mathcal{X}_N)$ \Comment{$\in [-1,1]^{C \times C}$}
    
    \State $\vPhi \gets \textsc{FillNaN}(\vPhi, 0)$
    
    \State $\vPhi_{\text{abs}} \gets |\vPhi|$ \Comment{Element-wise absolute value}
    
    \State ${f}_d \gets \{i \mid \vPhi_{\text{abs}}[i,:] = \mathbf{0} \}$ 
    \Comment{Channel index with zero correlation profile vector}
    
    \State ${f}_c \gets \{1,2,\dots,C\} \setminus f_d$ \Comment{Remaining channel index $f_c$, and \(C^{'} = |f_c|\))}
    
    \State $\vPhi^{'}_{\text{abs}} \gets \vPhi_{\text{abs}}[{f}_c, {f}_c]$ \Comment{$\in [0,1]^{C^{'} \times C^{'}}$}
    
    \State $\vW \gets \textsc{CosineSimilarity}(\vPhi^{'}_{\text{abs}})$ \Comment{ Using Eq. ~\eqref{eq:cosine_similarity}. $\vW\in [0,1]^{C^{'} \times C^{'}}$}
    
    \State $\vD_{ii} \gets \sum_j \vW_{ij}$ 

    \State $\vL \gets \vI - \vD^{-1/2} \vW \vD^{-1/2}$
    
    \If{$f_d = \emptyset$}
        \State $\vV \gets$ \(2^{nd}\) to $(M{+}1)^{th}$ smallest eigenvectors of $\vL$ \Comment{$\in \mathbb{R}^{C^{'} \times M}$}
        \State $\vU \gets \vD^{-1/2} \vV$  \Comment{$\in \mathbb{R}^{C^{'} \times M}$}
        \State $k^{'} \gets \textsc{K-Means}(\vU, M)$
    \Else
        \State $\vV \gets$ \(2^{nd}\) to $M^{th}$ smallest eigenvectors of $\vL$ \Comment{$\in \mathbb{R}^{C^{'} \times (M{-}1)}$}
        \State $\vU \gets \vD^{-1/2} \vV$  \Comment{$\in \mathbb{R}^{C^{'} \times (M-1)}$}
        \State $k^{'} \gets \textsc{K-Means}(\vU, M{-}1)$
    \EndIf
    
    \State $k \gets \text{zeros}(C)$
    \State $k[{f}_c] \gets k^{'}$, \quad $k[{f}_d] \gets M$
    \State \Return $k$
\EndFunction
\end{algorithmic}
\end{algorithm}

\section{Datasets}
\label{appendix_datasets}
We evaluate the effectiveness of our model using six publicly available datasets: SWaT, WADI, PSM, SMD, SMAP, and MSL. 
\begin{itemize}
    \item The SWaT (Secure Water Treatment)~\cite{swat} dataset contains time-series data from 51 sensors and actuators of a water treatment system. It spans 11 days, with the first 7 days representing normal operation and the remaining 4 days including attack scenarios.
    \item WADI (Water Distribution)~\cite{wadi} dataset includes data from 123 sensors and actuators over 16 days. The first 14 days correspond to normal operation, and the last 2 days involve attack scenarios.
    \item PSM (Pooled Server Metrics)~\cite{psm} data is collected from multiple application servers at eBay. This dataset comprises 25 dimensions, each representing a different server.
    \item SMD (Server Machine Dataset)~\cite{smd} is a multi-entity dataset containing telemetry data from 28 server machines. Each machine provides data from 38 sensors, and the duration of the dataset is five weeks.
    \item SMAP (Soil Moisture Active Passive) and MSL (Mars Science Laboratory)~\cite{smap, smap_msl} are multi-entity datasets provided by NASA. These consist of real spacecraft telemetry data, including recorded anomalies. SMAP contains 55 entities with 25 dimensions, while MSL includes 27 entities with 55 dimensions.
\end{itemize}
 The details of the datasets are shown in Table~\ref{tab:dataset} of the main paper.
\section{Data processing}
\label{data_processing}
Following the approach of SensitiveHUE~\cite{sensitivehue}, we downsampled the SWAT and WADI datasets by a factor of 5 to ensure fair comparisons. Data normalization was performed using min-max scaling, with parameters computed from the training set. To prevent extreme values from affecting the model, we clipped the normalized test data to the range \([-4, +4]\), as recommended by NPSR~\cite{npsr}. We adopted the preprocessing steps below following~\cite{npsr, sensitivehue}:
\begin{itemize}
    \item For the SWAT dataset, the P201 and LIT401 channels exhibited inconsistencies between the training and test sets; hence, we set their values to zero~\cite{sensitivehue}. To reduce noise, we applied a moving average filter to the entire dataset.
    \item For the WADI dataset (2017 version), we ignored columns with excessive missing values and forward-filled a small number of remaining NaNs. One channel was also set to zero based on the same reference. Additionally, we excluded the first 21,600 data points from the training set to account for system initialization~\cite{npsr}.
    \item For the PSM dataset, we applied a moving average for noise reduction, and forward-filled all missing values~\cite{npsr}.
\end{itemize}

\section{Baselines and Result Sources}
\label{appendix_baselines}
We employ \camready{23 baselines}, including autoencoder-based, recurrent-based, generative, graph-based, transformer-based, MLP-mixer-based, and
classical outlier detection approaches.\par
Specifically, DAGMM~\cite{dagmm}, UAE~\cite{uae}, and USAD~\cite{usad} are autoencoder-based methods, while LSTM-VAE~\cite{lstm_vae} adopts a variational autoencoder framework. MSCRED~\cite{mscred}, OmniAnomaly~\cite{smd}, and MAD-GAN~\cite{mad_gan} represent CNN-based, recurrent-based, and GAN-based approaches, respectively. GDN~\cite{gdn} and MTAD-GAT~\cite{mtad_gat} are graph-based models. We further include transformer-based methods such as Anomaly Transformer~\cite{anomaly_transformer}, D3R~\cite{d3r}, TranAD~\cite{tranad}, SAT~\cite{sat}, SimAD~\cite{simAD}, NPSR~\cite{npsr}, and SensitiveHUE~\cite{sensitivehue}. In addition, PatchAD~\cite{patchad} is an MLP-mixer-based model, while OracleAD~\cite{oracleAD} is an LSTM-based approach that explicitly models causality. Furthermore, we include PCA-Error~\cite{position}, a classical linear reconstruction-based baseline for anomaly detection. \camready{We have also included CAROTS~\cite{CAROTS}, a causality-aware contrastive learning method; CATCH~\cite{CATCH}, a frequency-domain patching-based reconstruction method; TimesNet~\cite{timesnet}, a 2D temporal variation modeling backbone; and xLSTMAD~\cite{xlstmad}, an encoder-decoder xLSTM-based anomaly detection method.} \par

To obtain the reported \textbf{best F1 scores}, we primarily refer to the original publications of each baseline. When the original paper does not report best-F1 scores, we adopt the corresponding results from subsequent peer-reviewed works. Since not all baselines evaluate on all datasets, some entries in~\cref{tab:main_results} are unavailable.\par
SensitiveHUE reports results only on the SWAT, WADI, MSL, and SMD datasets. For completeness, we evaluate SensitiveHUE (Offline) on the PSM and SMAP datasets, and SensitiveHUE (Online) on all datasets using the publicly available code. \camready{In addition, TimesNet reports point-adjusted F1, which has known flaws, and xLSTMAD does not report per-dataset F1; we therefore reproduce these baselines using their publicly available code for all datasets.} A detailed summary of the result sources for all baselines is provided in~\cref{tab:baseline_source}. 

\begin{table}[!htb]
\centering
\scriptsize
\caption{\camready{Details of the baselines and their corresponding result sources. The listed sources primarily refer to the reported best-F1 results. For additional metrics (e.g., PR AUC, VUS PR, and VUS ROC), the corresponding values, when available, are obtained from the same sources as the best-F1 scores.}}
\label{tab:baseline_source}
\begin{tabular}{@{}ccc@{}}
\toprule
Baselines & Venue & Result Source \\ \midrule
DAGMM~\cite{dagmm} & ICLR-2018 & NPSR \\
LSTM-VAE~\cite{lstm_vae} & IEEE-2018 & NPSR \\
MSCRED~\cite{mscred} & AAAI-2019 & NPSR \\
OmniAnomaly~\cite{smd} & KDD-2019 & NPSR \\
MAD-GAN~\cite{mad_gan} & ICANN-2019 & NPSR \\
MTAD-GAT~\cite{mtad_gat} & ICDM-2020 & NPSR \\
USAD~\cite{usad} & KDD-2020 & NPSR \\
UAE~\cite{uae} & IEEE-2020 & NPSR \\
GDN~\cite{gdn} & AAAI-2021 & NPSR \\
Anomaly Trans.~\cite{anomaly_transformer} & ICLR-2022 & NPSR \\ 
TimesNet~\cite{timesnet} & ICLR-2023 & Evaluated by us \\ 
D3R~\cite{d3r} & NeurIPS-2023 & SimAD \\ \bottomrule
\end{tabular}
\hspace{2pt}
\begin{tabular}{@{}ccc@{}}
\toprule
Baselines & Venue & Result Source \\ \midrule
TranAD~\cite{tranad} & VLDB-2022 & NPSR \\
NPSR~\cite{npsr} & NeurIPS-2023 & NPSR \\
SAT~\cite{sat} & IJCAI-2024 & SAT \\
OracleAD~\cite{oracleAD} & NeurIPS-2025 & OracleAD \\
PCA-Error~\cite{position} & ICML-2025 & Position \\
PatchAD~\cite{patchad} & IEEE-2025 & PatchAD \\
SimAD~\cite{simAD} & IEEE-2025 & SimAD \\
Sens.HUE(Offline)~\cite{sensitivehue} & KDD-2024 & SensitiveHUE \\
Sens.HUE(Online)~\cite{sensitivehue} & KDD-2024 & Evaluated by us \\ 
CAROTS~\cite{CAROTS} & ICML-2025 & CAROTS \\
CATCH~\cite{CATCH} & ICLR-2025 & CATCH \\
xLSTMAD~\cite{xlstmad} & IEEE-2025 & Evaluated by us \\ \bottomrule
\end{tabular}
\end{table}

\section{Definition of Best-F1 Score}
\label{definition_f1}
Our anomaly detection framework depends on the statistical significance $\alpha$ and the threshold $h$ (~\cref{anomaly_detction_method}). So, the predicted label $\hat{y}_t \in \{0, 1\}$ is a function of $\alpha$ and $h$. True Positive ($\mathrm{TP}$), False Positive ($\mathrm{FP}$), and False Negative ($\mathrm{FN}$)
are defined as follows:
\begin{align}
    \mathrm{TP} &= \{\, t \mid \hat{y}_t = 1,\; y_t = 1 \,\} \\
    \mathrm{FP} &= \{\, t \mid \hat{y}_t = 1,\; y_t = 0 \,\} \\
    \mathrm{FN} &= \{\, t \mid \hat{y}_t = 0,\; y_t = 1 \,\}
\end{align}
where $\hat{y}_t$ and $y_t$ denote the predicted and ground-truth labels at time step $t$,
respectively. The Precision ($\mathrm{P}$), Recall ($\mathrm{R}$), and F1 score are derived by
\begin{align}
    \mathrm{P} &= \frac{\#\mathrm{TP}}{\#\mathrm{TP} + \#\mathrm{FP}} \\
    \mathrm{R} &= \frac{\#\mathrm{TP}}{\#\mathrm{TP} + \#\mathrm{FN}} \\
    \mathrm{F1} &= \frac{2PR}{P + R}.
\end{align}

The Best-F1 score is obtained by selecting $\alpha$ and $h$ using the threshold sweeping method described in~\cite{sensitivehue, npsr} to maximize the F1 score:
\begin{equation}
    \mathrm{Best\text{-}F1} = \max_{\alpha,\, h} \; \mathrm{F1}\bigl(\hat{\vec{y}}(\alpha, h), \vec{y}\bigr),
\end{equation}
where $\hat{\vec{y}}$ and $\vec{y}$ are the set of predicted and ground truth labels.

\begin{table}[!htb]
\centering
\small
\caption{Search range of the hyperparameters}
\label{tab:hyperparameters}
\begin{tabular}{@{}r|c|c|c|c|c|c|@{}} \toprule
 & SWAT & WADI & PSM & SMAP & MSL & SMD \\ \midrule
Number of clusters \((M)\) & \([1,8]\) & \([1,10]\) & \([1,7]\) & \([1,10]\) & \([1,10]\) & \([1,10]\) \\ \cmidrule{2-7}
Initial learning rate & \multicolumn{6}{c|}{\([10^{-5}, 10^{-2}]\)} \\ \cmidrule{2-7}
Look-back window length \((L)\) & \multicolumn{6}{c|}{24} \\ \cmidrule{2-7}
\(\alpha\) (Equ. ~\ref{alpha}) & \([0, 0.001]\) & \([0, 0.047]\) & \([0, 0.603]\) &  \multicolumn{3}{c|}{[0, 1)} \\ \cmidrule{2-7}
Embedding mixer exp. factor \((d_f)\) & \multicolumn{6}{c|}{\{1, 3, 5\}} \\ \cmidrule{2-7}
Embedding dimension \((d)\) & \multicolumn{6}{c|}{\{128, 256\}} \\ \cmidrule{2-7}
Train epoch & \multicolumn{6}{c|}{30} \\ \cmidrule{2-7}
Batch size & \multicolumn{6}{c|}{512} \\ \bottomrule
\end{tabular}
\end{table}

\begin{figure*}[!htb]
  \centering
  \subfloat{\includegraphics[width=0.5\textwidth]{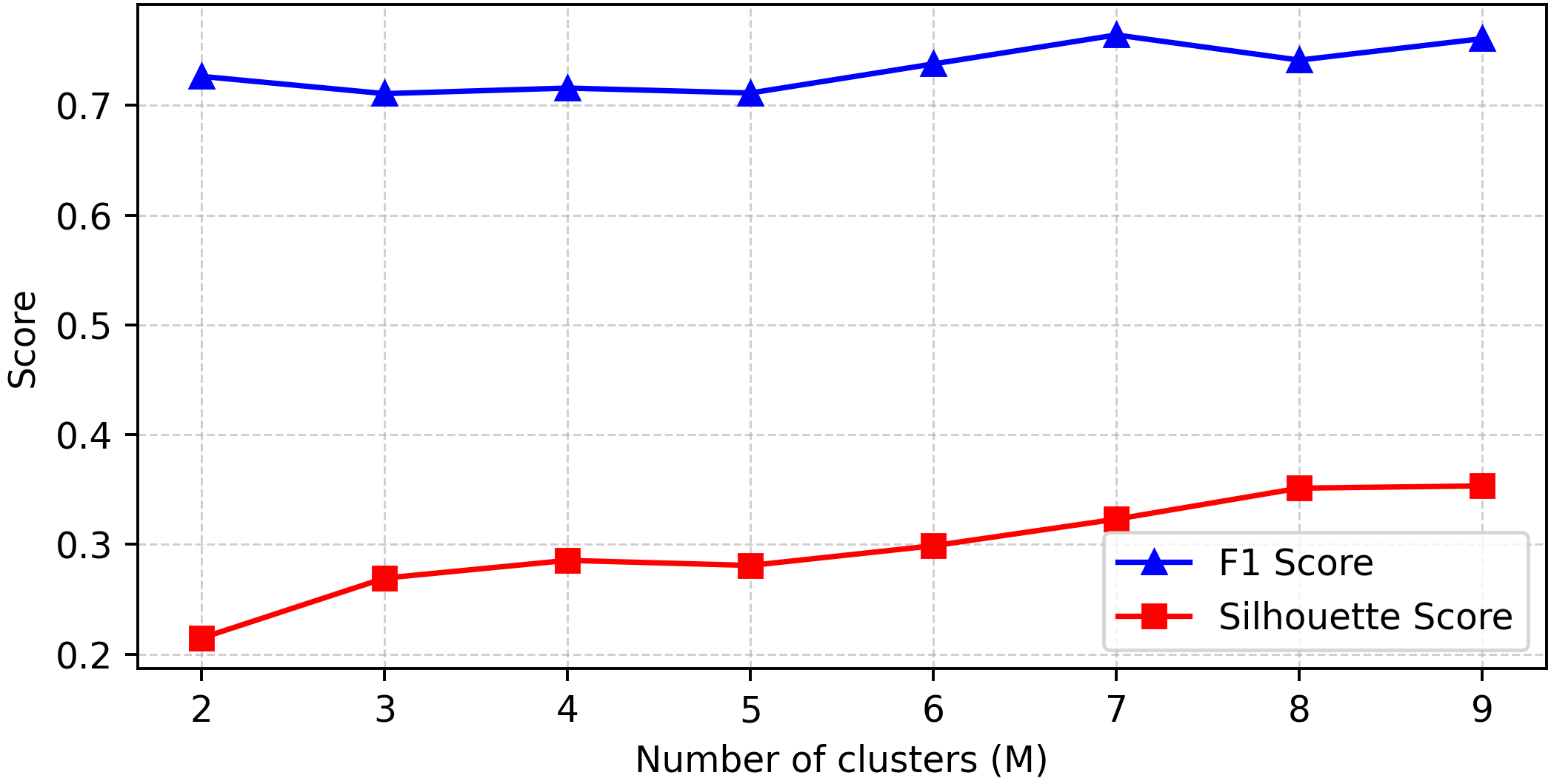}}
  \caption{Performance of the model on the WADI dataset with varying \( M \). A higher silhouette score indicates better clustering quality. Cluster numbers greater than 9 are ignored, as they result in at least one cluster containing a single channel.}
  \label{fig:M_sensitivity}
\end{figure*}

\section{Training and Hyperparameter Search}
\label{training_details}
We allocate 20\% of the training dataset for validation purposes. Experiments with multi-entity datasets are performed in three-protocols: (i) training a single model across all entities, (ii) training separate models per entity and averaging entity-wise F1 scores, and (iii) training per-entity models and aggregating predictions before computing F1. Unless otherwise specified, results for multi-entity datasets correspond to protocol (iii), following~\cite{sensitivehue}. Hyperparameter optimization is performed using the Optuna framework. The ranges of the hyperparameters searched during hyperparameter optimization are listed in Table~\ref{tab:hyperparameters}. For all experiments, we use the Adam optimizer to optimize the model. We also use MSE as the reconstruction loss to optimize the reconstruction model during training. We set the look-back window length to 24 and \(\delta\) to 5 for all experiments.\par

To determine the optimal number of clusters $M$, we first compute \( \phi_{\mathrm{abs}}\) from the training data, where each row represents the correlation profile of a channel. Channels are then clustered based on these profiles. To narrow the search space for $M$, we use the silhouette score as a measure of clustering quality, where a higher score indicates better separation~\cite{silhouette}. Since directly optimizing $M$ over a wide range of $M$ (e.g., \( M = 2 \) to \( 50 \) for a dataset with 100 channels) is computationally expensive, the silhouette score is used to preselect a smaller, promising range. As shown in ~\cref{fig:M_sensitivity}, the $F1$ score peaks within the same range of $M$ (from 6 to 9) that yields higher silhouette scores, confirming the effectiveness of silhouette-guided selection. The model is subsequently fine-tuned using Optuna across this range to determine the final $M$. \par

Our model is implemented in PyTorch 2.5.1 with CUDA 11.8, using Python 3.10.12. All experiments are conducted on a system equipped with an NVIDIA GeForce RTX 4090 GPU (24 GB), an AMD Ryzen 9 7950X 16-core processor, and 128 GB of RAM.\par

\subsection{Selection of the significance parameter $\alpha$}
\label{Selection_of_alpha}
\begin{figure*}[!tb]
  \centering
  \subfloat[PSM\label{fig:PSM_vali_loss_distribution}]{\includegraphics[width=0.33\textwidth]{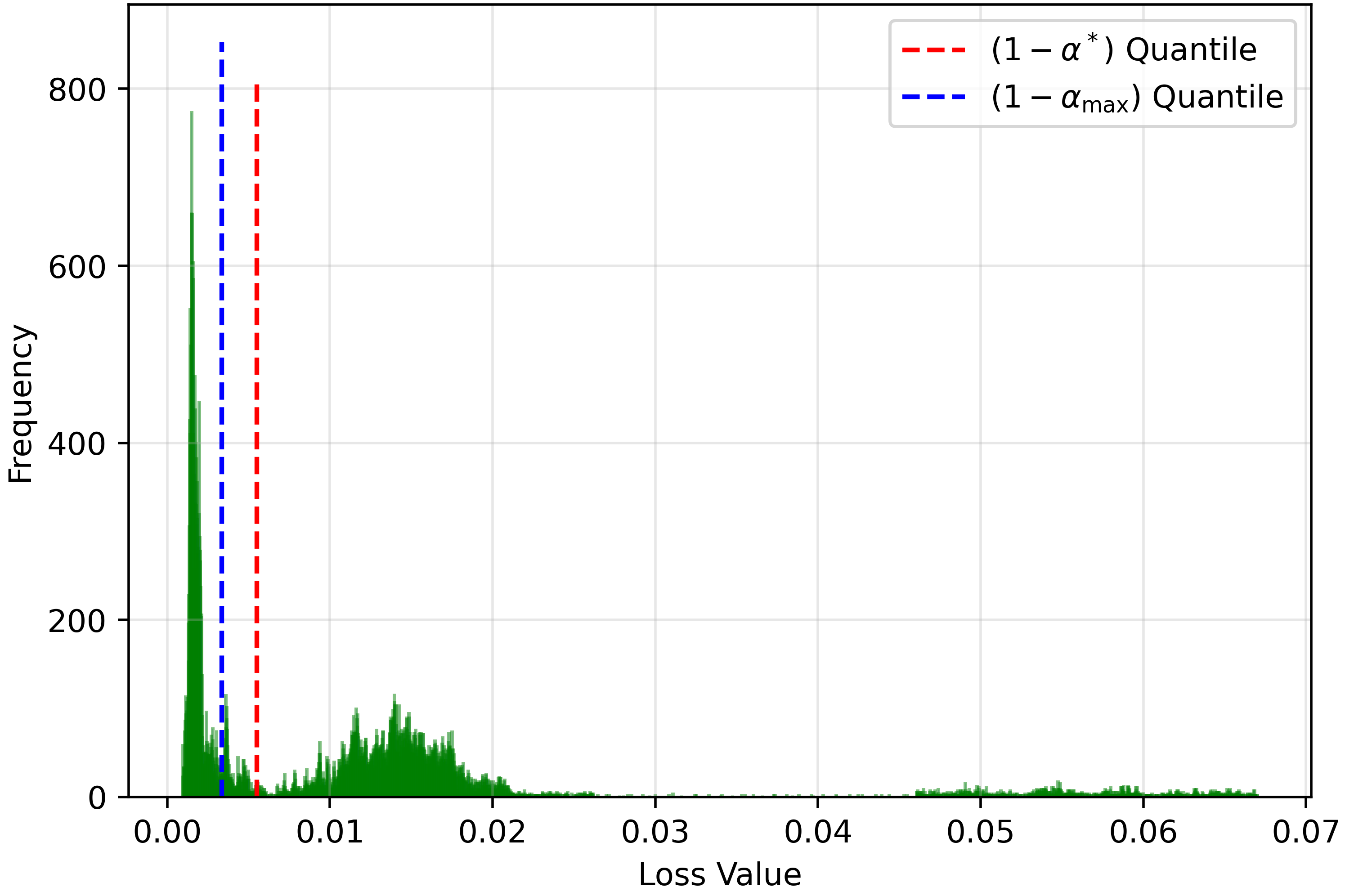}}\hfill
  \subfloat[WADI\label{fig:WADI_vali_loss_distribution}]{\includegraphics[width=0.33\textwidth]{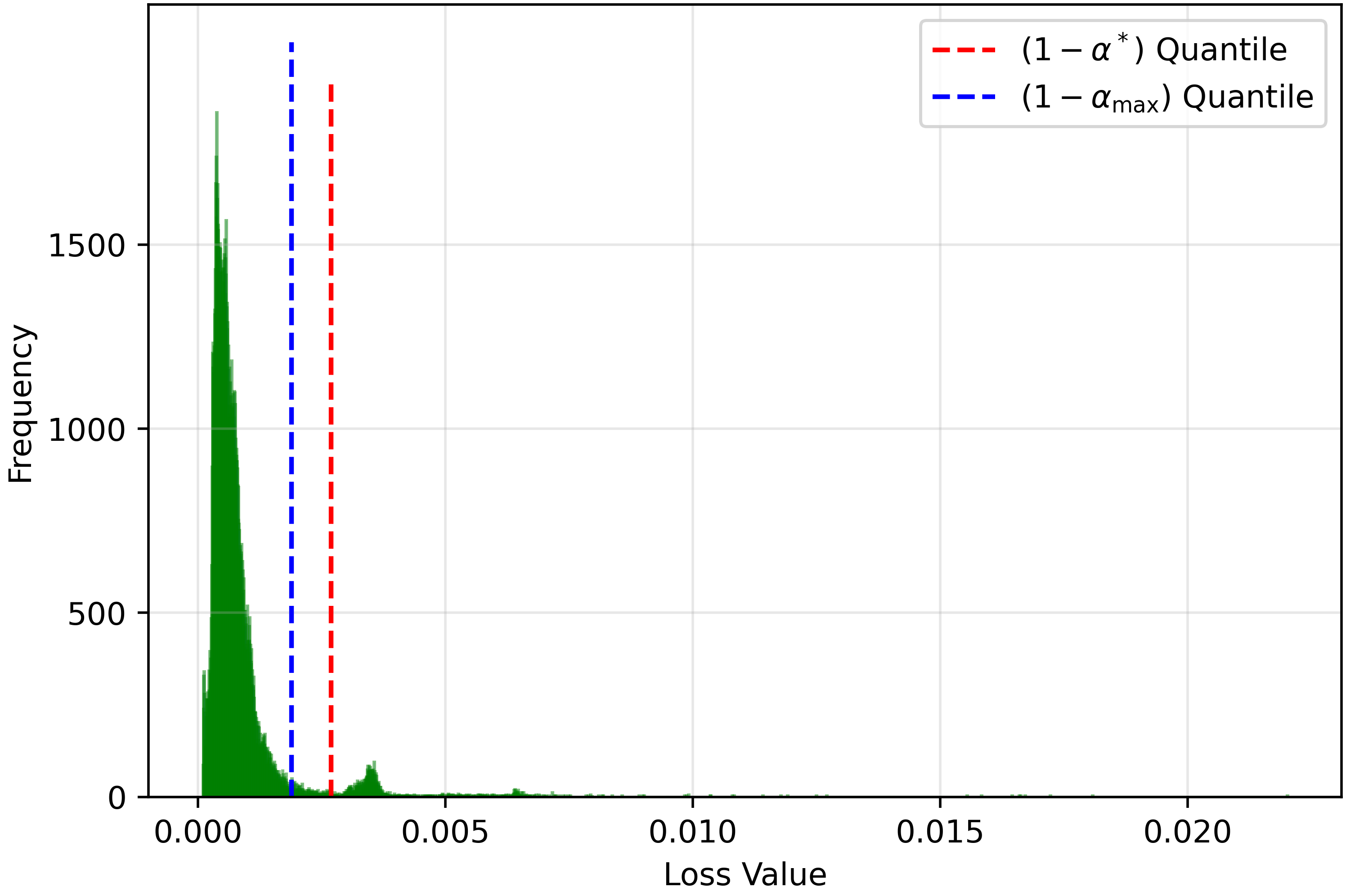}}\hfill
  \subfloat[SWAT\label{fig:SWAT_vali_loss_distribution}]{\includegraphics[width=0.33\textwidth]{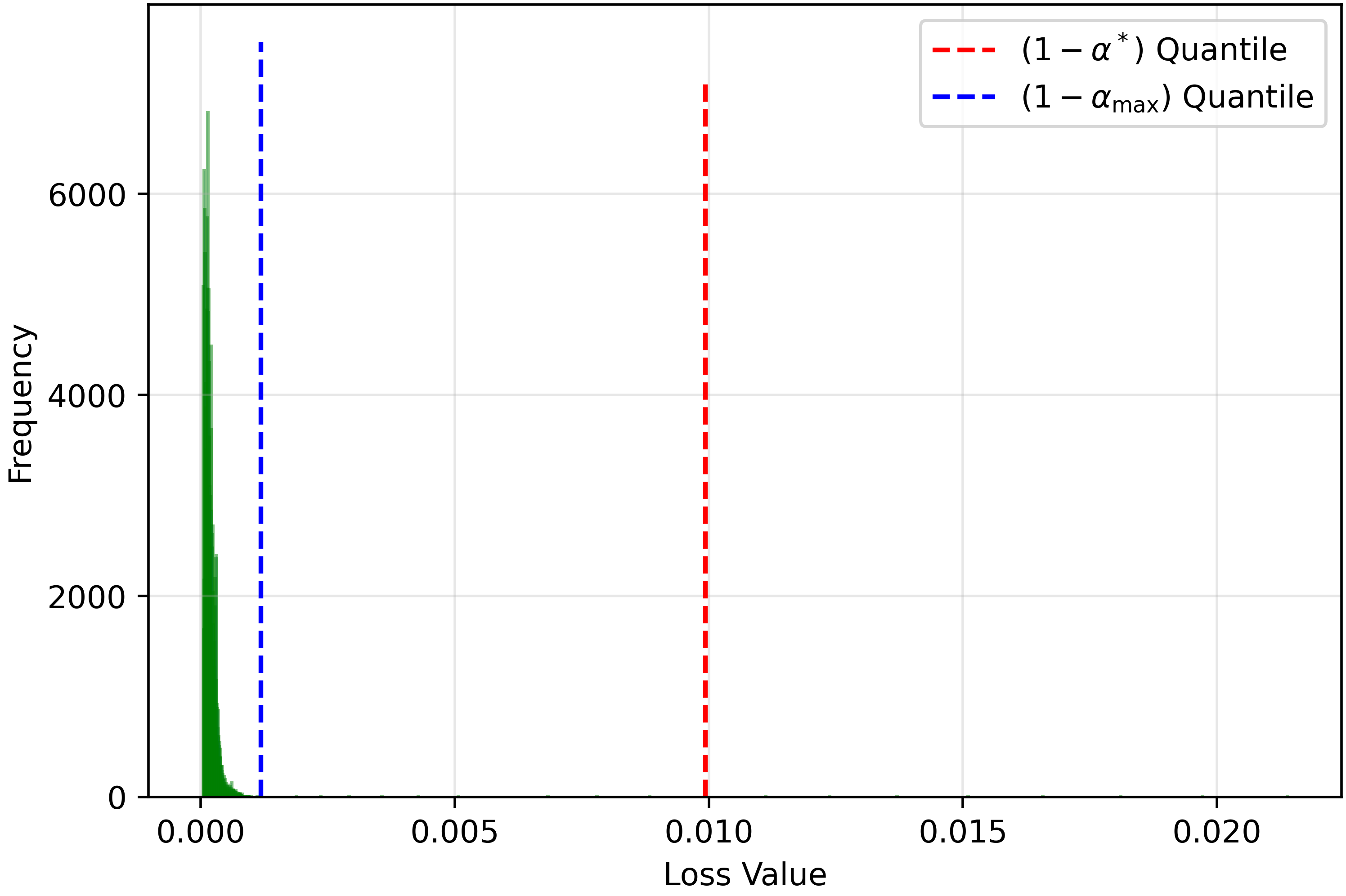}}\hfill
  
  \caption{Histograms of the validation loss distributions for different datasets. Here, $\alpha^{*}$ denotes the optimal value that yields the highest F1 score, while $\alpha_{\max}$ represents the maximum value of the search range for $\alpha$. The value of $\alpha_{\max}$ for the corresponding datasets are mentioned in~\cref{tab:hyperparameters}.}
  \label{fig:validation_loss_distribution}
\end{figure*}

The parameter $\alpha$ controls the statistical significance of anomaly evidence by defining a $(1-\alpha)$-percentile over the validation reconstruction loss distribution. For datasets with sufficiently large validation sets (SWaT, PSM, and WADI), we empirically observe that the validation reconstruction loss can be well characterized by a mixture of Gaussian components, where the dominant component with mean closest to zero. Based on this observation, we employ a principled heuristic to restrict the search range of $\alpha$. Specifically, we identify the Gaussian component whose mean is nearest to zero and compute an upper bound in the loss domain as $\mu + \sigma$ of that component. This value is then mapped to its empirical quantile to obtain $(1-\alpha_{\max})$, which defines the upper limit $\alpha_{\max}$. The optimal significance parameter $\alpha^{*}$ is subsequently searched within the reduced interval $[0, \alpha_{\max}]$, substantially narrowing the search space. As illustrated in Figure~\ref{fig:validation_loss_distribution}, the selected $\alpha^{*}$ consistently lies well below $\alpha_{\max}$, indicating that this bound serves as a conservative and effective range reduction rather than a hard threshold. \par

In contrast, for datasets with limited validation samples due to entity-wise training (MSL, SMAP, and SMD), reliable estimation of the validation loss distribution is not feasible. Therefore, for these datasets, $\alpha$ is searched over the full range $[0, 1)$ to avoid bias introduced by unstable distributional estimates.

\subsection{Training vs. Inference Behavior}
\label{training_vs_inf_behavior}
During training, Batch Normalization relies on batch-level statistics. However, during inference, the model operates in evaluation mode using fixed running statistics. As anomaly detection is performed only at inference time, predictions and anomaly scores at time $t$ do not depend on future observations, ensuring temporal causal behavior in detection.

\begin{table}[]
\footnotesize
\centering
\caption{\camready{Performance comparison using additional metrics (VUS\_PR, VUS\_ROC) across six benchmark datasets. For multi-entity datasets, baselines annotated with superscript$^{*i}$ correspond to results under evaluation protocol-$i$.}}
\label{tab:additional_results}

\begin{tabular}{@{}cccccc|cc@{}}
\toprule
Dataset & Metrics & TimesNet$^{*1}$ & xLSTMAD$^{*1}$ & CATCH$^{*1}$ & CCM-TAD$^{*1}$ & Sen.H.(On)$^{*2}$ & CCM-TAD$^{*2}$ \\ \midrule
\multirow{2}{*}{MSL}  & VUS\_PR & 0.211 & 0.180 & \textbf{0.256} & 0.249 & 0.291 & \textbf{0.581} \\
                      & VUS\_ROC & 0.703 & 0.646 & \textbf{0.735} & 0.660 & 0.673 & \textbf{0.777} \\ \cmidrule{2-6} \cmidrule{7-8}
\multirow{2}{*}{SMD}  & VUS\_PR & 0.220 & 0.108 & 0.159 & \textbf{0.263} & 0.349 & \textbf{0.492} \\
                      & VUS\_ROC & 0.795 & 0.620 & \textbf{0.797} & 0.525 & \textbf{0.801} & 0.750 \\ \cmidrule{2-6} \cmidrule{7-8}
\multirow{2}{*}{SMAP} & VUS\_PR & 0.120 & 0.145 & 0.155 & \textbf{0.326} & 0.335 & \textbf{0.533} \\
                      & VUS\_ROC & 0.471 & 0.527 & 0.543 & \textbf{0.569} & 0.700 & \textbf{0.740} \\ \bottomrule
\end{tabular}

\vspace{8pt}

\begin{tabular}{@{}ccccccc@{}}
\toprule
Dataset & Metrics & TimesNet & xLSTMAD & CATCH & Sen.H.(On) & CCM-TAD \\ \midrule
\multirow{2}{*}{SWAT}  & VUS\_PR & 0.223 & 0.469 & 0.241 & 0.729 & \textbf{0.779} \\
                       & VUS\_ROC & 0.541 & 0.625 & 0.462 & \textbf{0.828} & 0.758 \\ \cmidrule{2-7}
\multirow{2}{*}{PSM}   & VUS\_PR & 0.428 & 0.484 & 0.436 & 0.454 & \textbf{0.485} \\
                       & VUS\_ROC & 0.617 & 0.621 & 0.639 & \textbf{0.676} & 0.545 \\ \cmidrule{2-7}
\multirow{2}{*}{WADI}  & VUS\_PR & 0.379 & 0.550 & - & 0.597 & \textbf{0.776} \\
                       & VUS\_ROC & 0.813 & \textbf{0.861} & - & 0.857 & 0.822 \\ \bottomrule
\end{tabular}

\end{table}

\begin{table}[]
\footnotesize
\centering
\caption{\camready{Performance comparison between CCM-TAD and CATCH using Aff\_F1 and R\_F1 metrics.}}
\label{tab:results_aff_range_f1}

\begin{tabular}{@{}ccccccc@{}}
\toprule
Metrics                   & Models   & PSM   & SWAT  & MSL$^{*1}$   & SMD$^{*1}$   & SMAP$^{*1}$ \\ \midrule

\multirow{2}{*}{R\_F1}    & CCM-TAD  & 0.428 & \textbf{0.681} & \textbf{0.292} & \textbf{0.293} & \textbf{0.347} \\
                          & CATCH    & \textbf{0.498} & 0.148 & 0.185 & 0.158 & 0.173 \\ \midrule
                          
\multirow{2}{*}{Aff\_F1}  & CCM-TAD  & 0.694 & \textbf{0.781} & \textbf{0.756} & 0.695 & \textbf{0.703} \\
                          & CATCH    & \textbf{0.859} & 0.755 & 0.740 & \textbf{0.847} & 0.699 \\ \bottomrule
\end{tabular}
\end{table}

\section{Additional Results}
\label{Additional_results}
In addition to the best F1 score, we present our model's performance using VUS\_ROC and VUS\_PR, which are shown in~\cref{tab:additional_results}. The VUS metrics were introduced to mitigate excessive penalization caused by minor boundary misalignment between detected anomaly regions and ground-truth annotations, which are often unavoidable in time-series data~\cite{paparrizos2022volume}. \camready{The results show that our model consistently achieves strong performance in VUS\_PR across most datasets, outperforming the baselines included in the table in the majority of cases. Although the improvements in VUS\_ROC are comparatively less consistent, VUS\_PR is generally more informative than VUS\_ROC for highly imbalanced anomaly detection tasks~\cite{saito2015precision}. Additionally, in Table~\ref{tab:results_aff_range_f1}, we compare our model against CATCH using Aff\_F1 and R\_F1. It shows that our model also outperforms CATCH in these two metrics in most cases.}

\begin{figure*}[!htb]
  \centering
  \includegraphics[width=0.5\textwidth]{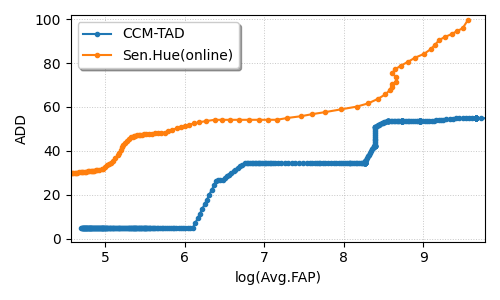}
  \caption{ADD-FAP Trade-off comparison for WADI dataset.}
  \label{fig:add_fap}
\end{figure*}
Figure~\ref{fig:add_fap} shows the trade-off between average detection delay (ADD) and the logarithm of the average false-alarm period (log(Avg.FAP)), which is commonly used to evaluate the performance sequential change/anomaly detection methods. False positive rate (i.e., false alarm probability) corresponds to the inverse of FAP. Our method (CCM-TAD) attains lower ADD than SensitiveHUE(online) at comparable false alarm periods, indicating faster detection for the same false-alarm period. 
Overall, the CCM-TAD curve lies consistently below the SensitiveHUE curve, demonstrating a more favorable ADD-FAP trade-off for online anomaly detection.

\section{Computational Cost and Robustness}
\label{computational_cost}
Table~\ref{tab:robustness} reports the robustness comparison between our model and SensitiveHUE (Online) across four metrics, with results averaged over five random seeds. Our model achieves superior performance on most metrics across the datasets. While SensitiveHUE occasionally attains higher VUS-ROC scores, this metric is less informative under severe class imbalance. In anomaly detection settings, VUS-PR is more appropriate, as it better reflects precision–recall trade-offs and penalizes false positives, consistent with prior studies favoring PR-based metrics over ROC-based metrics for imbalanced data~\cite{saito2015precision}.\par
Table~\ref{tab:appendix_computational_cost} reports the computational cost in terms of GFLOPs per batch, number of trainable parameters, and average inference time. Compared to SensitiveHUE (Online), our model consistently requires fewer GFLOPs across all datasets. In terms of model size, CCM-TAD uses fewer parameters on SWAT, PSM, WADI, and SMAP, while requiring more parameters on MSL and SMD. Although CCM-TAD incurs higher inference latency across datasets compared to SensitiveHUE (Online), the absolute inference times remain low (on the order of milliseconds per batch). Overall, these results indicate that CCM-TAD improves robustness and detection performance with a moderate and controlled computational overhead.

\begin{table}[!htb]
\footnotesize
\centering
\caption{Robustness Comparison}
\label{tab:robustness}
\begin{tabular}{@{}clcc@{}}
\toprule
Dataset & Metrics & CCM-TAD (ours) & Sen.HUE(Online) \\ \midrule
\multirow{4}{*}{SWAT} & F1 & 0.877 ± 0.004 & 0.850 ± 0.033 \\
 & PR\_AUC & 0.907 ± 0.004 & 0.832 ± 0.022 \\
 & VUS\_ROC & 0.749 ± 0.011 & 0.819 ± 0.014 \\
 & VUS\_PR & 0.770 ± 0.009 & 0.711 ± 0.025 \\ \cmidrule{2-4}
\multirow{4}{*}{PSM} & F1 & 0.688 ± 0.022 & 0.502 ± 0.003 \\
 & PR\_AUC & 0.558 ± 0.023 & 0.457 ± 0.004 \\
 & VUS\_ROC & 0.546 ± 0.007 & 0.668 ± 0.005 \\
 & VUS\_PR & 0.487 ± 0.013 & 0.449 ± 0.003 \\ \cmidrule{2-4}
\multirow{4}{*}{WADI} & F1 & 0.744 ± 0.015 & 0.679 ± 0.006 \\
 & PR\_AUC & 0.698 ± 0.024 & 0.617 ± 0.004 \\
 & VUS\_ROC & 0.813 ± 0.017 & 0.864 ± 0.011 \\
 & VUS\_PR & 0.756 ± 0.016 & 0.598 ± 0.010 \\ \bottomrule 
\end{tabular}
\hspace{3mm}
\begin{tabular}{@{}clcc@{}}
\toprule
Dataset & Metrics & CCM-TAD (ours) & Sen.HUE(Online) \\ \midrule
\multirow{4}{*}{MSL} & F1 & 0.633 ± 0.023 & 0.421 ± 0.005 \\
 & PR\_AUC & 0.592 ± 0.024 & 0.280 ± 0.006 \\
 & VUS\_ROC & 0.782 ± 0.015 & 0.670 ± 0.005 \\
 & VUS\_PR & 0.564 ± 0.024 & 0.291 ± 0.006 \\ \cmidrule{2-4}
\multirow{4}{*}{SMD} & F1 & 0.507 ± 0.068 & 0.336 ± 0.019 \\
 & PR\_AUC & 0.561 ± 0.030 & 0.404 ± 0.006 \\
 & VUS\_ROC & 0.744 ± 0.010 & 0.799 ± 0.004 \\
 & VUS\_PR & 0.460 ± 0.032 & 0.347 ± 0.005 \\ \cmidrule{2-4}
\multirow{4}{*}{SMAP} & F1 & 0.558 ± 0.030 & 0.532 ± 0.014 \\
 & PR\_AUC & 0.571 ± 0.020 & 0.346 ± 0.003 \\
 & VUS\_ROC & 0.743 ± 0.007 & 0.699 ± 0.006 \\
 & VUS\_PR & 0.551 ± 0.016 & 0.334 ± 0.002 \\ \bottomrule 
\end{tabular}

\end{table}

\begin{table}[!htb]
\centering
\small
\caption{Comparison of computational costs in terms of GFLOPs per batch, number of trainable parameters, and average inference time per batch. For multi-entity datasets, the reported values are averaged over all entities. All experiments are performed on the same hardware described in~\cref{training_details}.}
\label{tab:appendix_computational_cost}

\begin{tabular}{@{}c|ccc|ccc|ccc@{}}
\toprule
 & \multicolumn{3}{c|}{SWAT} & \multicolumn{3}{c|}{PSM} & \multicolumn{3}{c}{WADI} \\ \midrule
 & GFLOPs & \#Param. & Inf. time 
 & GFLOPs & \#Param. & Inf. time 
 & GFLOPs & \#Param. & Inf. time \\ \midrule
Sen.HUE(Online) & 3.73 & 155k & 0.0008 & 13.41 & 551k & 0.0008 & 28.19 & 1152k & 0.0015 \\
CCM-TAD & 1.05 & 43k & 0.0015 & 3.55 & 144k & 0.002 & 10.68 & 435k & 0.0021 \\ \bottomrule
\end{tabular}
\par\vspace*{6pt}
\begin{tabular}{@{}c|ccc|ccc|ccc@{}}
\toprule
 & \multicolumn{3}{c|}{MSL} & \multicolumn{3}{c|}{SMD} & \multicolumn{3}{c}{SMAP} \\ \midrule
 & GFLOPs & \#Param. & Inf. time 
 & GFLOPs & \#Param. & Inf. time 
 & GFLOPs & \#Param. & Inf. time \\ \midrule
Sen.HUE(Online) & 14.02 & 291k & 0.0014 & 13.7 & 284k & 0.0013 & 26.34 & 1077k & 0.0013 \\
CCM-TAD & 10.6 & 432k & 0.0028 & 8.98 & 366k & 0.0022 & 10.47 & 426k & 0.0021 \\ \bottomrule
\end{tabular}

\end{table}

\section{Additional Ablation Studies}
\label{Additional_Ablation_Studies}

\subsection{Channel Clustering Variants}
\label{ablation_additional_channel_clustering_approaches}
To evaluate the effectiveness of our proposed spectral clustering method based on correlation profiles, we conduct a comprehensive ablation study comparing three alternative channel clustering strategies. These strategies isolate the influence of different similarity metrics and clustering paradigms on the structural organization of features and their downstream impact on anomaly detection performance. \par
\paragraph{(1) Channel Similarity-Based Spectral Clustering:}
In this variant, we construct the similarity-based adjacency matrix \(\vW \in \mathbb{R}^{C \times C}\) using the cosine similarity between channel time series. The similarity score \(w_{ij} \in [0, 1]\) is computed as,
\begin{align}
    w_{ij} =
        \begin{cases}
        \max\left(0,\frac{\langle \mathcal{X}_{N(:,i)}, \mathcal{X}_{N(:,j)} \rangle}{\left\| \mathcal{X}_{N(:,i)} \right\| \left\| \mathcal{X}_{N(:,j)} \right\|}\right), & \text{if } i \neq j \\
        0, & \text{if } i = j
        \end{cases}
\end{align}
where, \(\mathcal{X}_{N(:,i)} \in \mathbb{R}^{N \times 1}\) is the \(i^{th}\) channel of the training data \(\mathcal{X}_N \in \mathbb{R}^{N \times C}\). \(N\) and \(C\) are the length and the number of channels of the training data, respectively. We then apply a similar spectral clustering approach described in Section~\ref{feature_clustering} to the resulting similarity weight matrix.
\paragraph{(2) Absolute Correlation-Based Spectral Clustering:}
In this configuration, the adjacency matrix \(\vW \in \mathbb{R}^{C \times C}\) is directly derived from the absolute Pearson correlation matrix \(\vPhi_{\text{abs}} \in \mathbb{R}^{C \times C}\), where \(w_{ij} = \phi_{ij}\). Spectral clustering is then performed on this graph to determine channel groupings. This method captures the pairwise correlation without considering the full correlation profile.
\paragraph{(3) K-Means Clustering on Correlation Matrix:}
Here, instead of forming a graph and applying spectral clustering, we directly cluster the rows of \(\vPhi_{\text{abs}}\) using the K-Means algorithm. Each row \(\phi_i\) can be interpreted as the correlation profile of the \(i\)th channel, representing its relationship with all other channels. This method evaluates the efficacy of clustering in the original correlation profile space, independent of the properties of the spectral graph. \par
The anomaly detection results of all these approaches are shown in Table~\ref{tab:different_clustering_methods} of the main paper. Our proposed approach, which utilizes spectral clustering on the full correlation profiles, consistently outperforms these baselines, highlighting the importance of preserving inter-channel relationships during clustering.

\begin{table}[!htb]
\footnotesize
\centering
\caption{Spurious correlation values obtained with clustering ($SC^W$) and without clustering ($SC^{W/O}$) for the PSM dataset. The column $\Delta SC (\%)$ denotes the percentage difference, computed as $\Delta SC (\%) = 100 \times (SC^{W/O} - SC^W) / SC^{W/O}$.}
\label{tab:ablation_spurious_correlation}
\begin{tabular}{@{}cccc@{}}
\toprule
Channel & $SC^W$ & $SC^{W/O}$ & $\Delta SC(\%)$  \\ \midrule
1 & 0.132 & 0.189 & 30.2 \\
2 & 0.177 & 0.253 & 30 \\
3 & 0.260 & 0.247 & -5.3 \\
4 & 0.225 & 0.285 & 21.1 \\
5 & 0.081 & 0.137 & 40.9 \\
6 & 0.098 & 0.151 & 35.1 \\
7 & 0.142 & 0.174 & 18.4 \\
8 & 0.076 & 0.166 & 54.2 \\
9 & 0.086 & 0.115 & 25.2 \\
10 & 0.085 & 0.124 & 31.5 \\
11 & 0.080 & 0.117 & 31.6 \\
12 & 0.092 & 0.162 & 43.2 \\
13 & 0.120 & 0.436 & 72.5 \\
14 & 0.126 & 0.557 & 77.4 \\
15 & 0.127 & 0.180 & 29.4 \\
16 & 0.155 & 0.173 & 10.4 \\
17 & 0.087 & 0.118 & 26.3 \\
18 & 0.085 & 0.143 & 40.6 \\
19 & 0.122 & 0.145 & 15.9 \\
20 & 0.079 & 0.070 & -12.9 \\
21 & 0.148 & 0.386 & 61.7 \\
22 & 0.123 & 0.187 & 34.2 \\
23 & 0.185 & 0.124 & -49.2 \\
24 & 0.277 & 0.290 & 4.5 \\
25 & 0.099 & 0.147 & 32.7 \\ \midrule
Avg. &  &  & 27.98 \\ \bottomrule
\end{tabular}
\end{table}

\subsection{Effect of Channel Clustering on Reducing Spurious Correlation in Prediction}
\label{ablation_channel_clustering}
\begin{figure*}[!htb]
  \centering
  \subfloat[SWAT\label{fig:SWAT_spurious_corr}]{\includegraphics[width=0.8\textwidth]{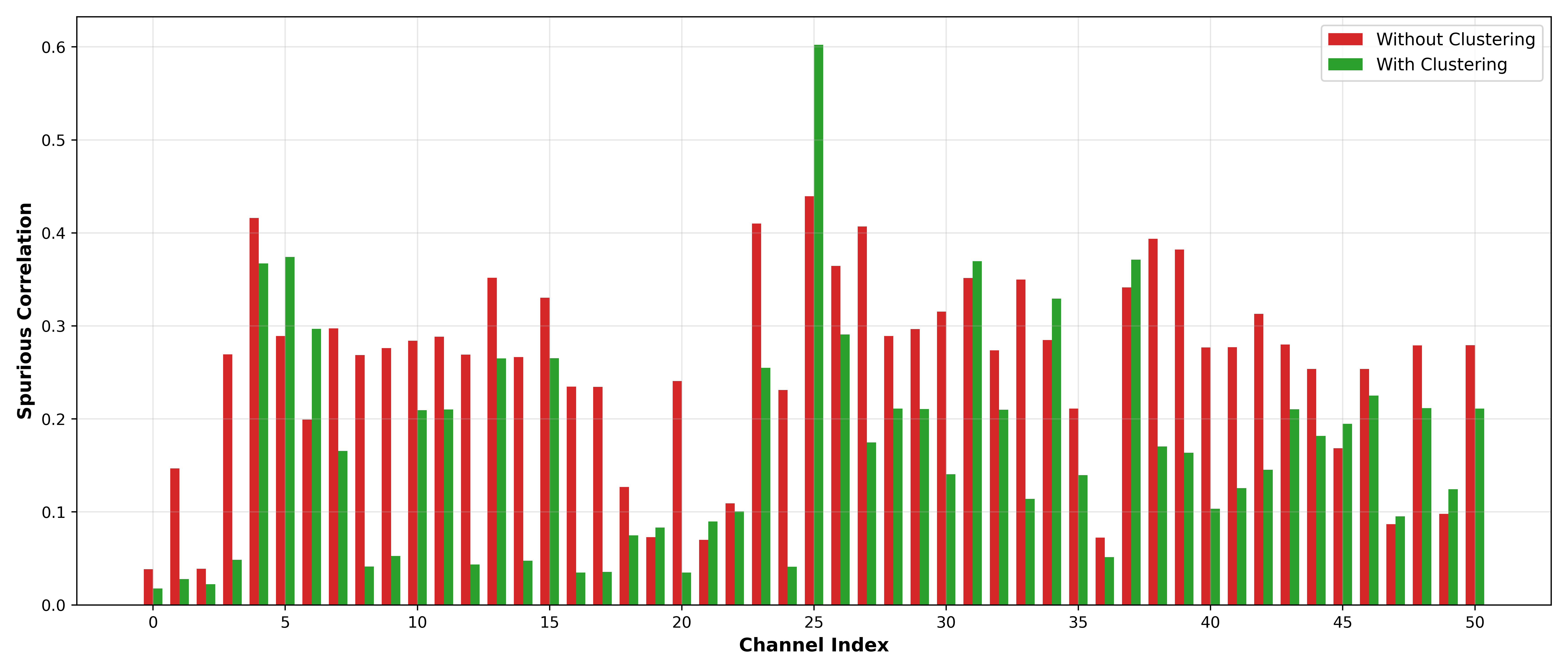}}\vfill
  \subfloat[WADI\label{fig:WADI_spurious_corr}]{\includegraphics[width=0.8\textwidth]{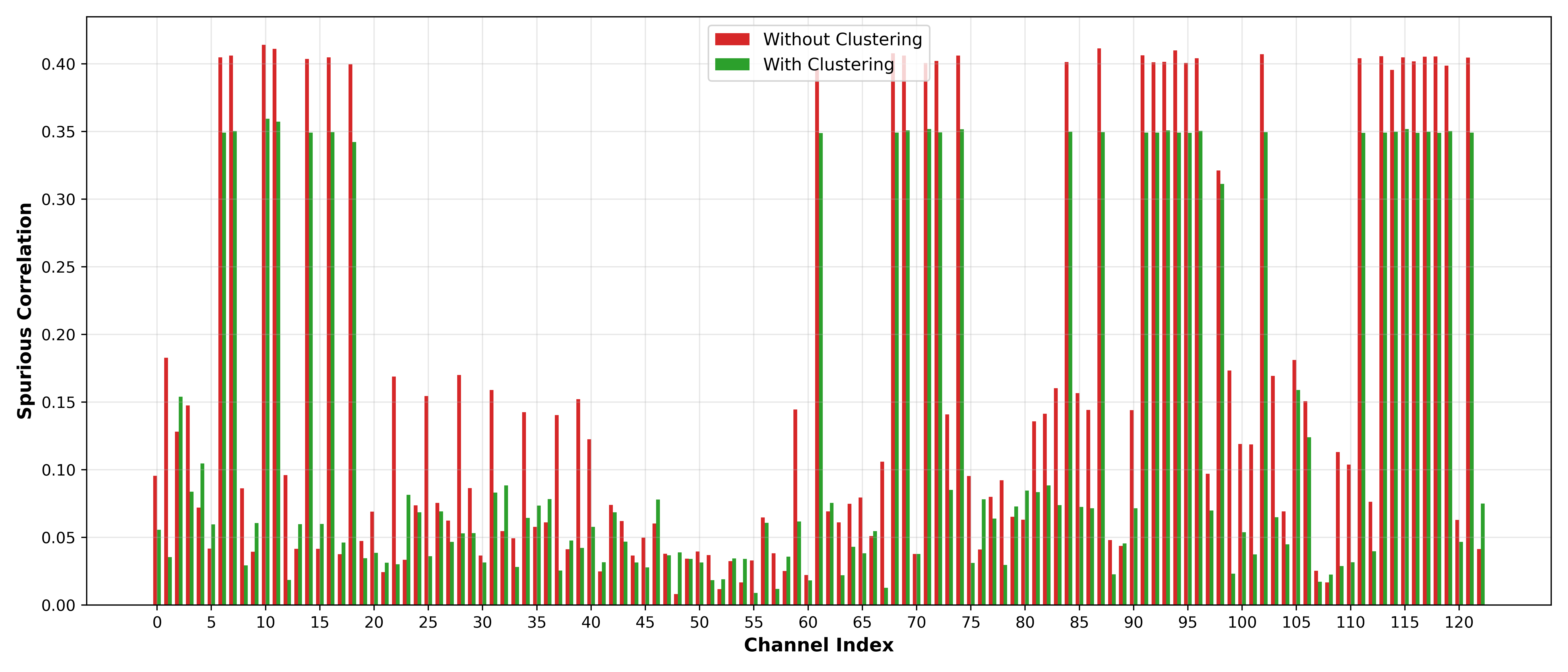}} 
  \caption{Spurious correlation reduction using our clustering approach}
  \label{fig:spurious_corr_reduc}
\end{figure*}
Our proposed model estimates channel correlations for normal data more accurately than the non-clustered approach. To evaluate the accuracy of the estimated correlations, we use the spurious correlation (SC) metric. The spurious correlation for channel-$i$ ($SC_i$) is defined as,
\begin{equation}
\mathrm{SC}_i = \frac{1}{C} \sum_{j=1}^{C} \left| \phi_{ij} - \hat{\phi}_{ij} \right|
\end{equation}
where $\phi_{ij}$ and $\hat{\phi}_{ij}$ denote the true and estimated correlations between channels $i$ and $j$, respectively, computed from the normal test data. Table~\ref{tab:ablation_spurious_correlation} presents the detailed results for the PSM dataset, showing that our clustering approach reduces the average spurious correlation among channels by 27.98\%. Similarly, the reductions for the SWAT and WADI datasets are 32.74\% and 14.22\%, respectively. Figure~\ref{fig:spurious_corr_reduc} further illustrates the spurious correlation reduction across the SWAT and WADI datasets.

\begin{figure}[!htb]
    \centering
    \begin{minipage}{0.49\textwidth}
        \centering
        \includegraphics[width=0.99\columnwidth]{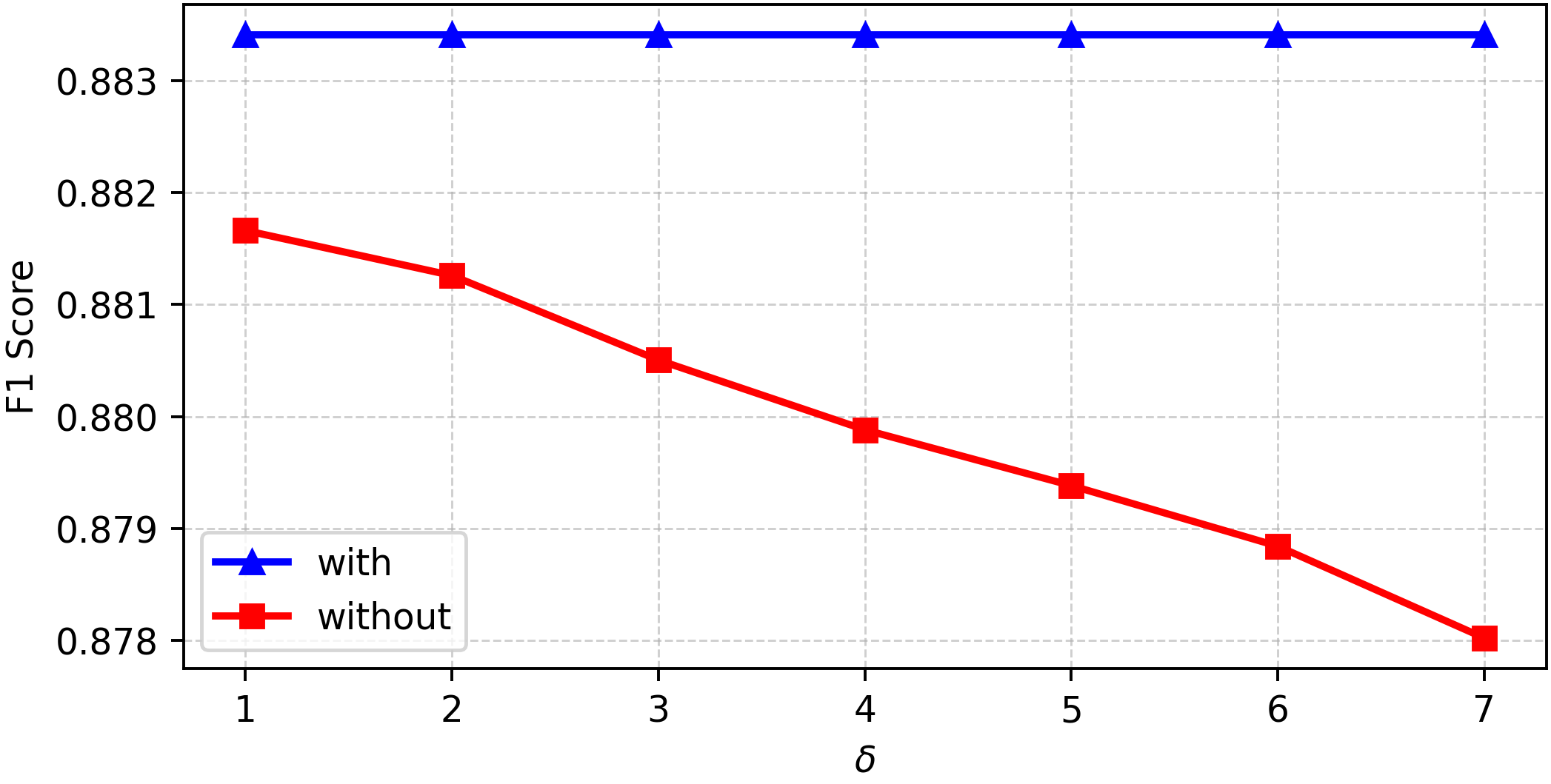}
        \caption{Performance of our proposed anomaly detection method for the SWAT dataset with varying $\delta$ is shown. ``With" denotes the method incorporating the optimal end-point detection formula~\cref{eq:anomaly_offset}, while ``Without" indicates its exclusion. The results demonstrate that performance remains consistent across different $\delta$ values when the optimal end-point detection is applied.}
        \label{fig:delta_sensitivity}
    \end{minipage}
    \hfill
    \begin{minipage}{0.49\textwidth}
        \centering
        \includegraphics[width=0.99\columnwidth]{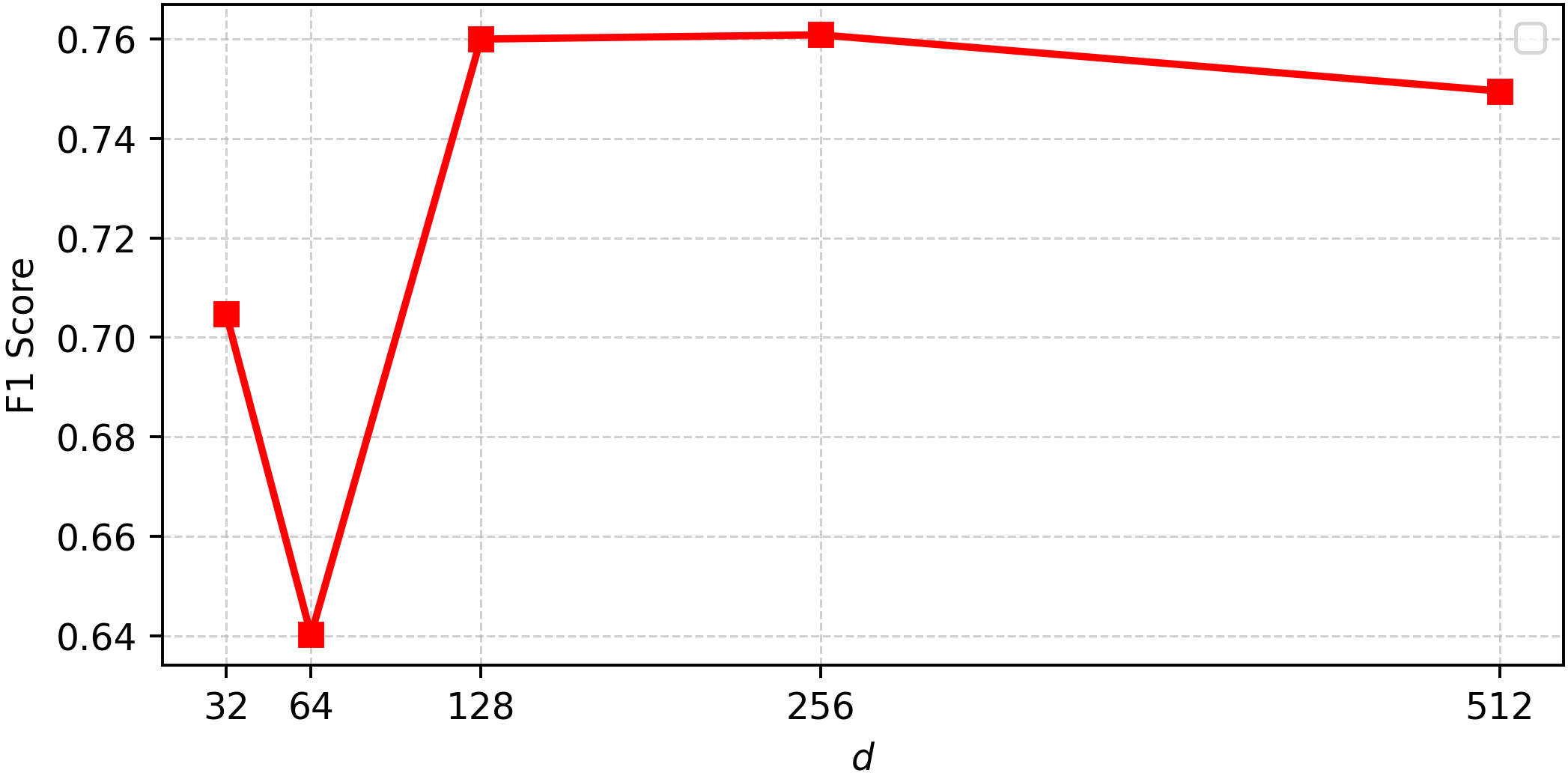}
        \caption{Performance of our model for the WADI dataset with varying $d$ is shown.}
        \label{fig:d_sensitivity}
    \end{minipage}
\end{figure}

\subsection{Sensitivity Analysis for the Hyperparameters}
\label{ablation_sensitivity_analysis}
\paragraph{Sensitivity Analysis on $\delta$:}
The parameter $\delta$ controls the decay of the accumulated anomaly evidence $s_t$ by requiring $\delta$ consecutive negative anomaly evidences $\beta_t$ before resetting $s_t$ to zero. Figure~\ref{fig:delta_sensitivity} shows that the anomaly detection performance remains largely unchanged across different $\delta$ settings. This robustness is attributed to the proposed anomaly boundary refinement mechanism~\cref{eq:anomaly_offset}, which determines the optimal anomaly endpoint of an anomaly sequence. 
\paragraph{Sensitivity Analysis on $d$:}
Figure~\cref{fig:d_sensitivity} illustrates the model’s performance on the WADI dataset for different values of $d\in \{32, 64, 128, 256, 512\}$. The F1 score varies with $d$ but shows no clear linear trend. Therefore, we determine the optimal $d$ using Optuna in our experiments.

\paragraph{Sensitivity Analysis for $\alpha$ and $h$:}
\label{sensitivity_analysis}
\begin{figure}[!tb]
  \subfloat[Sensitivity on $\alpha$\label{fig:alpha_vs_f1}]{\includegraphics[width=0.33\columnwidth]{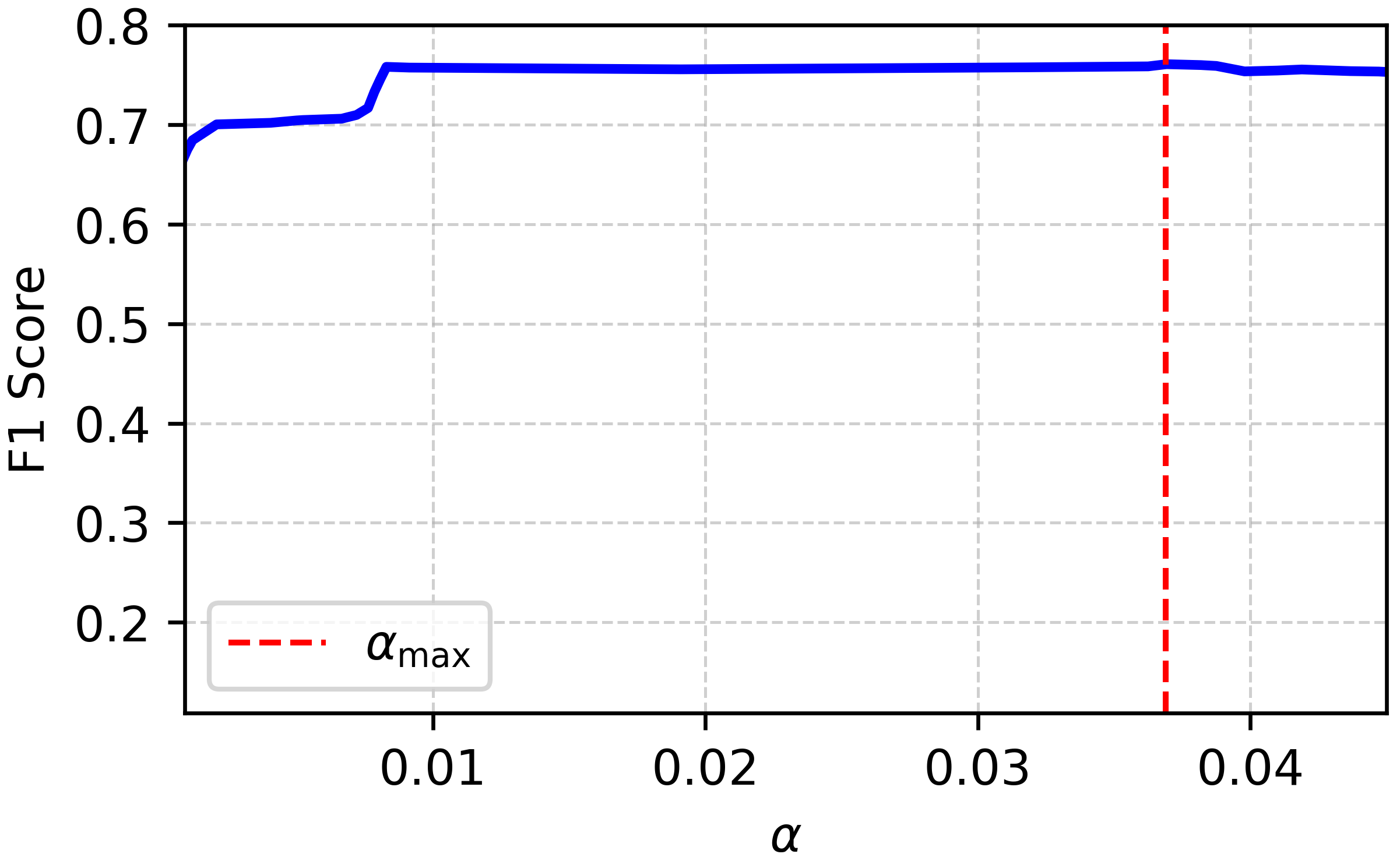}}\hfill
  \subfloat[Sensitivity on $h$\label{fig:h_vs_f1}]{\includegraphics[width=0.33\columnwidth]{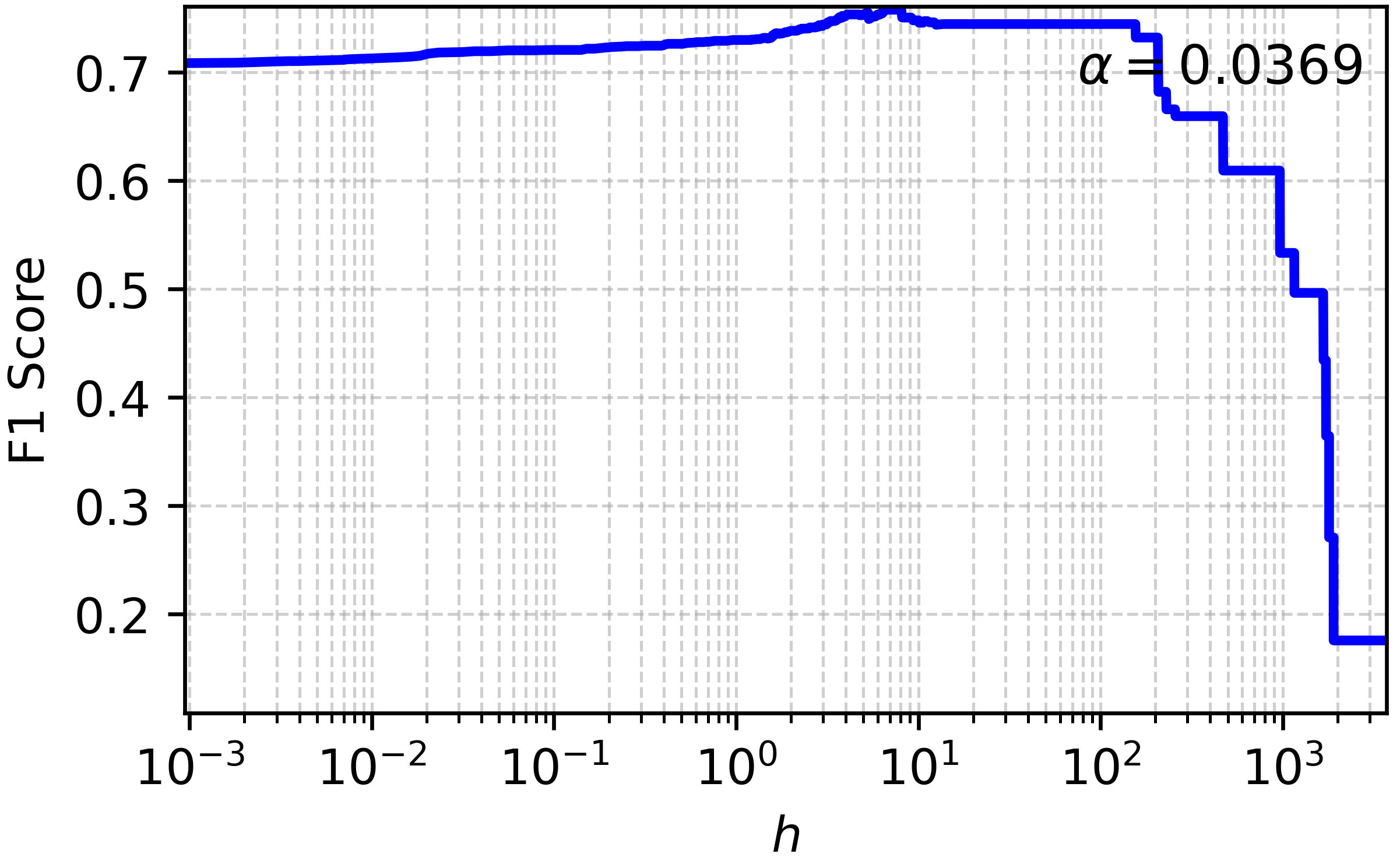}} \hfill
  \subfloat[Joint sensitivity on $\alpha$ and $h$\label{fig:joint_alpha_h_vs_f1}]{\includegraphics[width=0.33\columnwidth]{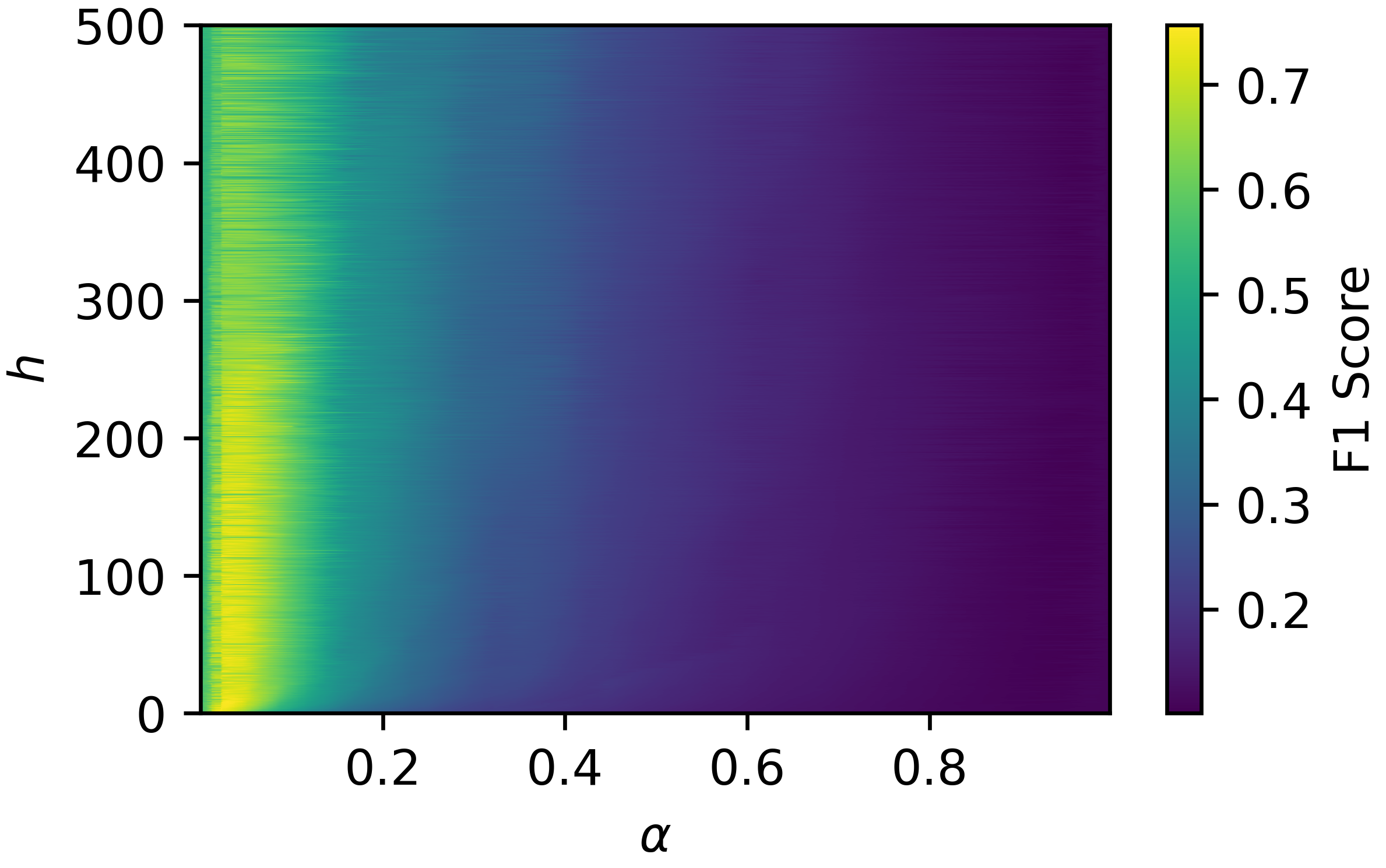}}
  \caption{F1 score sensitivity on $\alpha$ and $h$ for the WADI dataset.}
  \label{fig:alpha_h_sensitivity}
\end{figure}

The parameter $\alpha$ controls the statistical significance of anomaly evidence. Reconstruction error $g_t$ greater than the $(1-\alpha)$-percentile in validation set result in a positive anomaly evidence $\beta_t$ (Eq. \eqref{alpha}). Tuning this hyper-parameter allows the detector to adapt to dataset-specific nominal behavior. Figure~\ref{fig:alpha_vs_f1} shows the sensitivity of the F1 score on $\alpha$ for the WADI dataset, where smaller values of $\alpha$ achieve superior performance, indicating close alignment between validation and test normal losses. \camready{It also shows that the F1 score remains stable and consistently high within the relevant search range $[0, \alpha_{\max}]$, confirming that the method is robust within its operating region}. Figure~\ref{fig:h_vs_f1} analyzes the effect of the decision threshold $h$ for a fixed $\alpha$, showing stable performance up to $h \approx 10^2$, beyond which the F1 score degrades. Figure~\ref{fig:joint_alpha_h_vs_f1} presents the joint sensitivity on $\alpha$ and $h$, revealing a broad region of robust performance.

\subsection{Effect of Causal Mask's $1/j$ Scaling}
\label{ablation_1_over_j_mask}
\begin{figure}[!htb]
  \centering
  \subfloat[SWAT\label{fig:SWAT_mask_weight}]{\includegraphics[width=0.49\textwidth]{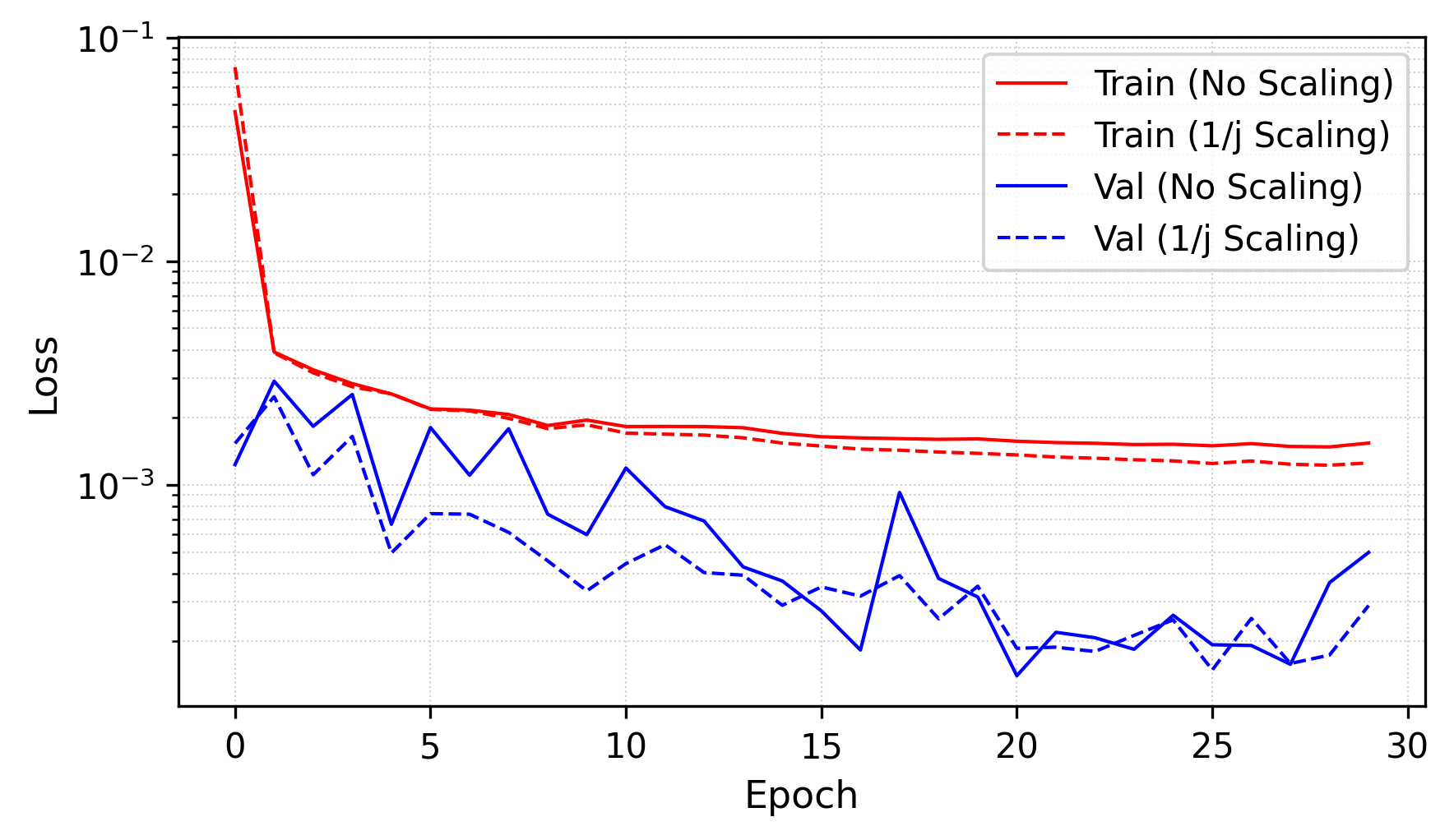}}\hfill
  \subfloat[WADI\label{fig:WADI_mask_weight}]{\includegraphics[width=0.49\textwidth]{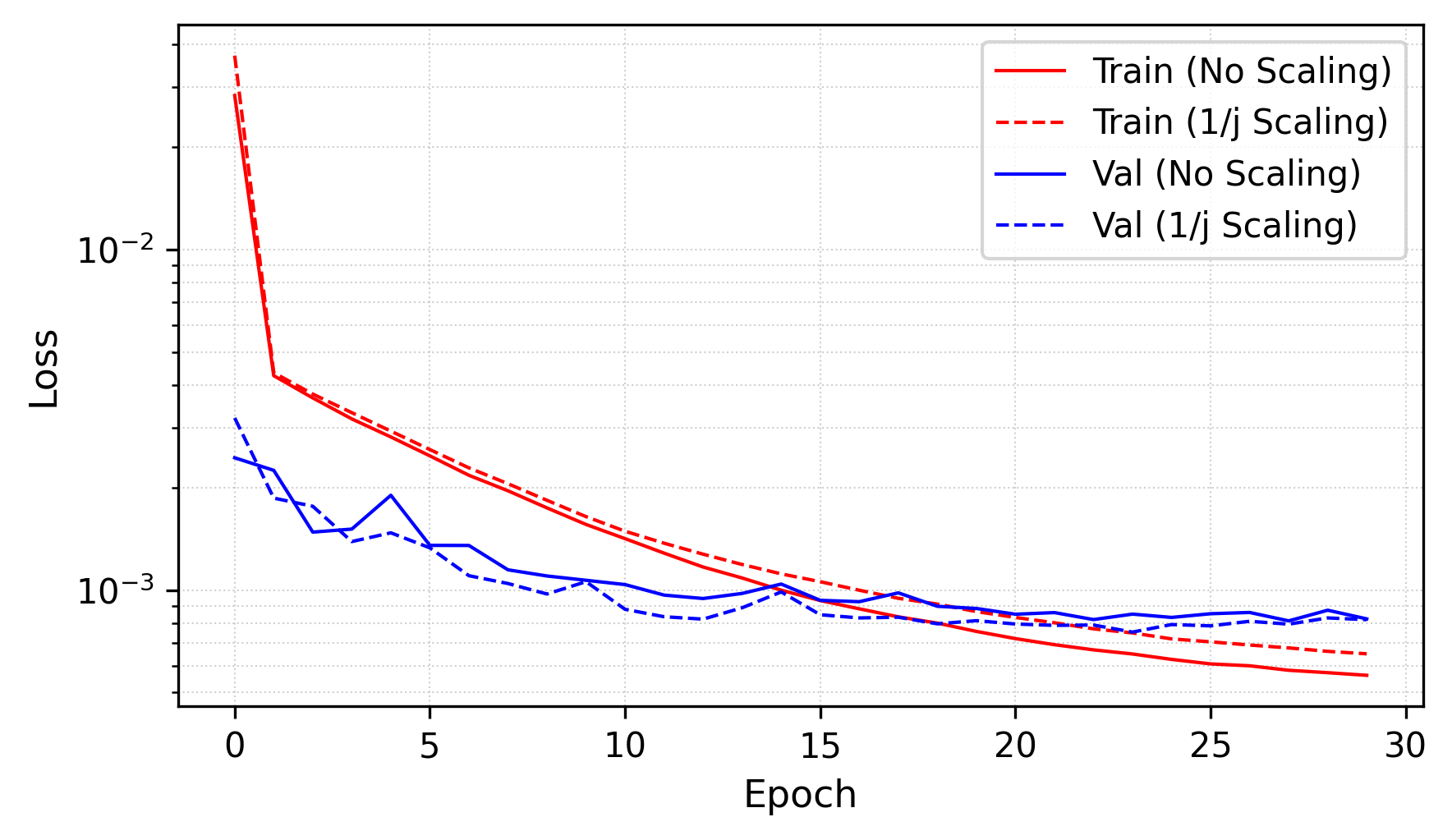}}
  \caption{Training and validation loss with and without $1/j$ scaling in the causal mask. ``No Scaling'' refers to unit masking ($\gamma_{ij}=1; i\leq j$).}
  \label{fig:masking_weight}
\end{figure}

Figure~\ref{fig:masking_weight} shows the effect of $1/j$ scaling in the causal temporal mask on the SWAT and WADI datasets. Here, ``No scaling'' refers to unit masking. While both unscaled and scaled masked variants exhibit similar convergence in training loss, the unscaled masked variant leads to noticeable oscillations and spikes in validation loss on the SWAT dataset. The WADI dataset also exhibits more oscillations in the validation loss with the unscaled masked variant.
Overall, these results suggest that incorporating $1/j$ scaling moderates the optimization dynamics.

\subsection{Impact of Distributional Shifts in Inter-Channel Correlation}
\label{appendxi_ablation_distributional_shift_correlation}

\begin{figure}[!htb]
  \centering
  \subfloat[SWAT\label{fig:SWAT_disti_corr}]{\includegraphics[width=0.9\textwidth]{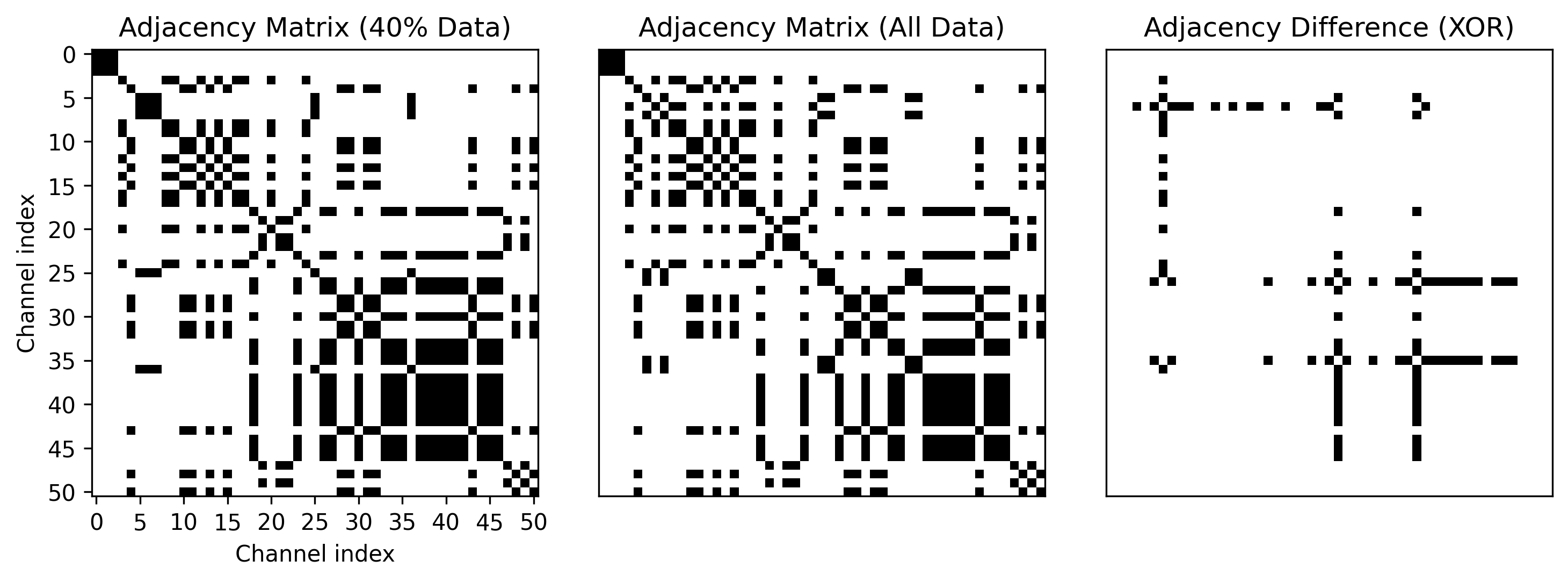}}\vfill
  \vspace{3pt}
  \subfloat[WADI\label{fig:WADI_disti_corr}]{\includegraphics[width=0.9\textwidth]{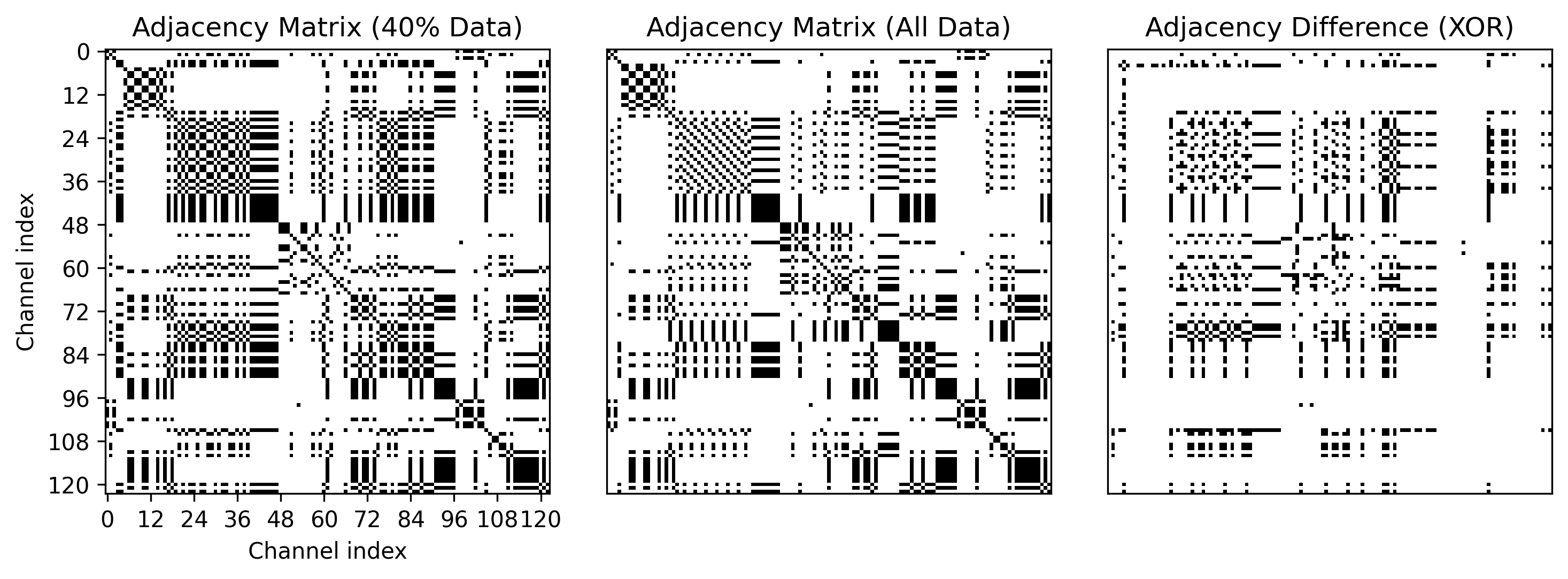}}
  \caption{Adjacency matrices presenting the graph structure of the channels, derived from different portions of the train dataset. Black and white entries denote the presence (1) and absence (0) of edges. The right panel shows the element-wise XOR between the two adjacency matrices, highlighting structural differences induced by data availability.}
  \label{fig:distributional_shift}
\end{figure}

Time-series data can exhibit distributional shifts in inter-channel correlations. To examine their impact, we consider two clustering settings: one using the first 40\% of the training data to derive channel clusters, and another using the full training set. The number of clusters is kept fixed, and the model is trained for both settings. As shown in \cref{fig:distributional_shift}, the adjacency difference (XOR) highlights changes in the induced graph structure caused by correlation shifts. Despite these changes, performance remains largely consistent. The best-F1 scores on SWAT are 0.878 and 0.883, while on WADI they are 0.762 and 0.761 for the two settings, respectively. This suggests robustness to moderate correlation shifts, while more substantial shifts may require re-estimation of the cluster numbers.

\section{Motivation for the Causal Mixer}
\label{motivation_causal_mixer}
In time series reconstruction, preserving temporal causality is critical. Without it, the representation at any time may be contaminated by future anomalies or vice versa. This can cause information leakage and distort the anomaly score. A conventional temporal mixer mixes information across the sequence without maintaining temporal order, which conflicts with the requirements of anomaly detection. We therefore mask the temporal mixer's weights to maintain temporal causality during information mixing. This directional information flow ensures effective model learning. \camready{Note that this causality constraint applies specifically to the information mixing process inside the temporal mixer, rather than to the last reconstructed point of the input window, as no future information is accessible at the last point of the input window.}

\section{Online Applicability of our Anomaly Detection Method}
\label{online_anomaly_detect_method}
Our proposed sequential anomaly detection method is applicable to detecting anomalies in real time. Unlike several prior methods, we do not apply any normalization to the \emph{anomaly score} that requires access to future test data. For example, SensitiveHUE normalizes anomaly scores using the median and interquartile range computed from the \emph{entire test set}, which necessitates observing all test samples beforehand and therefore limits its applicability to offline settings. In this paper, we denote their original implementation as ``SensitiveHUE(Offline)''.\par
In our method, the anomaly score $s_t$ at time $t$ depends solely on information available up to that time: (i) the reconstruction error at $t$, (ii) past anomaly evidences $\beta_{<t}$, (iii) past anomaly scores $s_{<t}$, and (iv) statistics derived from the validation set, which are obtained prior to deployment. No future test observations are required at any stage. Therefore, it operates in real-time.\par
Furthermore, the inference latency of our reconstruction model is low, as reported in Table~\ref{tab:appendix_computational_cost}, enabling timely processing of incoming measurements. These properties make the proposed approach suitable for practical online anomaly monitoring.

\begin{figure*}[!htb]
  \centering
  \subfloat{\includegraphics[width=0.65\textwidth]{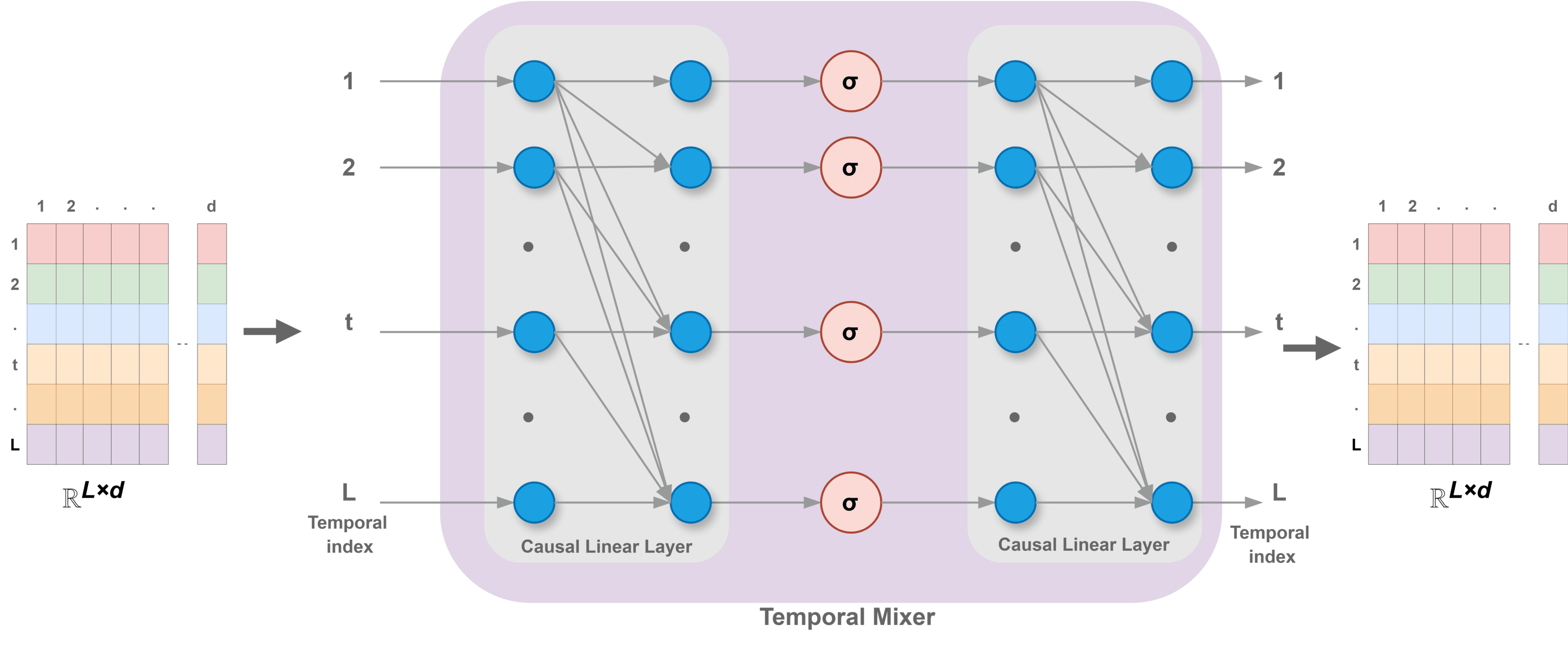}}
  \caption{Neural visualization of Temporal mixer module.}
  \label{fig:neural_temporal_mixer}
\end{figure*}

\section{Limitations}
\label{limitations}
Causality in time series analysis is a broad research area. This work focuses exclusively on \emph{temporal causality}. We do not aim to discover Granger causality, inter-channel causality, or the specific channels responsible for an anomaly~\cite{granger1}. In the temporal mixer, each neuron corresponding to a temporal index $t$ is allowed to interact only with neurons corresponding to indices $\le t$ from the previous layer, illustrated in~\cref{fig:neural_temporal_mixer}. This enforces directional information flow through weight masking, similar to the attention mask in transformers, but it does not model explicit causal graphs among variables. \camready{Moreover, since the model is built for multivariate time series anomaly detection, and cluster-aware multi-embedding by nature requires multiple channels, evaluating it on univariate datasets falls outside the intended scope of this work.}


\end{document}